\documentclass[10pt,journal,compsoc]{IEEEtran}
\usepackage{algorithm}
\usepackage{algorithmic}
\usepackage{csquotes} 
\usepackage{color}
\usepackage{paralist}
\usepackage{latexsym}
\usepackage{indentfirst}
\usepackage{subfigure}
\usepackage{float}
\usepackage{cite}
\usepackage{CJKutf8}
\usepackage{multirow}
\usepackage{makecell}
\usepackage{mathrsfs}
\usepackage[colorlinks, linkcolor=red,  anchorcolor=blue, citecolor=blue]{hyperref}
\usepackage{textcomp,booktabs}
\usepackage{amssymb}
\usepackage{pifont}
\newcommand{\cmark}{\ding{51}}%
\newcommand{\xmark}{\ding{55}}%
\usepackage{epsfig}
\usepackage{colortbl}
\usepackage{amsmath} 
\usepackage{siunitx}
\usepackage{longtable}
\definecolor{mygray}{gray}{.9}
\definecolor{mypink}{rgb}{.99,.91,.95}
\definecolor{mycyan}{cmyk}{.3,0,0,0}
\hypersetup{%
        anchorcolor=blue, 
        colorlinks,
        breaklinks=true,
        plainpages=false,%
        citecolor=blue,
        linkcolor=red,
        urlcolor=magenta,
        bookmarksopen=true,%
        bookmarksnumbered=false,%
        bookmarksdepth=5%
}

\usepackage{booktabs} 
\usepackage{makecell} 
\usepackage{lipsum} 

\begin{document}
\begin{CJK}{UTF8}{gbsn}


\title{Event Stream-based Visual Object Tracking: HDETrack V2 and A High-Definition Benchmark}  



\author{Shiao Wang, Xiao Wang*, \textit{Member, IEEE}, Chao Wang, Liye Jin, \\
    Lin Zhu, Bo Jiang*, Yonghong Tian, \textit{Fellow, IEEE}, Jin Tang 

\IEEEcompsocitemizethanks{
\IEEEcompsocthanksitem Shiao Wang, Xiao Wang, Chao Wang, Liye Jin, Bo Jiang, and Jin Tang are with the School of Computer Science and Technology, Anhui University, Hefei 230601, China. 
(email: \{wsa1943230570, w853023886\}@126.com, 18912870836@163.com, \{xiaowang, jiangbo, tangjin, luobin\}@ahu.edu.cn) 
\IEEEcompsocthanksitem Lin Zhu is with Beijing Institute of Technology, Beijing, China (email: linzhu@pku.edu.cn)
\IEEEcompsocthanksitem Yonghong Tian is with Peng Cheng Laboratory, Shenzhen, China; National Key Laboratory for Multimedia Information Processing, School of Computer Science, Peking University, China; School of Electronic and Computer Engineering, Shenzhen Graduate School, Peking University, China. (email: yhtian@pku.edu.cn) 
\IEEEcompsocthanksitem Corresponding author: Xiao Wang and Bo Jiang
}}

\markboth{IEEE Transactions on ******}%
{Shell \MakeLowercase{\textit{et al.}}: Bare Demo of IEEEtran.cls for Computer Society Journals}

\IEEEtitleabstractindextext{%
\begin{abstract} 
Recent years have witnessed unprecedented growth in the field of visual object tracking using bio-inspired event cameras. Previous research in this domain has primarily followed two paths: aligning RGB data and event streams for precise tracking or developing trackers that rely solely on events. However, while the former approach often incurs significant computational costs during inference, the latter may struggle with challenges posed by noisy event data or limited spatial resolution. In this paper, we present a hierarchical knowledge distillation framework designed to leverage multi-modal and multi-view information during the training phase, while utilizing only event signals for tracking. Specifically, we first train a teacher Transformer-based multi-modal/multi-view tracking network by simultaneously feeding RGB frames and event streams. We then introduce a novel hierarchical knowledge distillation strategy that incorporates the similarity matrix, feature representation, and response map-based distillation to guide the learning of the student Transformer network. We also enhance the model’s ability to capture temporal dependencies by applying the temporal Fourier transform to establish temporal relationships between video frames. We adapt the network model to specific target objects during testing via a newly proposed test-time tuning strategy to achieve high performance and flexibility in target tracking. Recognizing the limitations of existing event-based tracking datasets, which are predominantly low-resolution ($346 \times 260$), we propose EventVOT, the first large-scale high-resolution ($1280 \times 720$) event-based tracking dataset. It comprises 1141 videos spanning diverse categories such as pedestrians, vehicles, UAVs, ping pong, etc. Extensive experiments on both low-resolution (FE240hz, VisEvent, FELT), and our newly proposed high-resolution EventVOT dataset fully validated the effectiveness of our proposed method. Both the benchmark dataset and source code have been released on \url{https://github.com/Event-AHU/EventVOT_Benchmark}. 
\end{abstract}

}

\maketitle

\IEEEdisplaynontitleabstractindextext

%
\IEEEpeerreviewmaketitle

\IEEEraisesectionheading{\section{Introduction}\label{sec:introduction}}
\IEEEPARstart{V}{isual} object tracking~\cite{bertinetto2016staple, bhat2018unveiling, fan2019siamese, goutam2019Dimp, martin2019Atom, bolme2010visual, cao2023towards, cui2023mixformer, xie2024correlation, guo2024divert, li2024textureless}, has long been a fundamental task in the field of computer vision, which aims to locate the target object in video frames based on its appearance features initialized in the first frame. Existing trackers are primarily designed for RGB cameras and are widely deployed in autonomous driving, drone photography, intelligent video surveillance, and other fields. However, RGB cameras often struggle in extreme conditions such as overexposure, low lighting, and fast motion. These complex scenarios can occur in individual frames or even throughout an entire video, complicating efforts to enhance overall tracking performance through the mere addition of labeled data.

In recent years, biologically inspired event cameras have gradually entered people's field of vision. Different from traditional RGB cameras, event cameras continuously generate event signals by asynchronously capturing changes in lighting intensity within a scene. Concurrently, event cameras present a range of benefits over traditional RGB cameras, such as enhanced dynamic range, superior temporal resolution, reduced power requirements, and enhanced privacy preservation. It performs better in extreme conditions, such as low light and fast motion, and can be effectively applied to a range of tasks, including human action recognition~\cite{wang2024hardvs}, object detection~\cite{li2023sodformer}, and visual object tracking~\cite{wang2024event, wang2024visevent, huang2024mambafetrack}.

\begin{figure*}
\center
\includegraphics[width=6.5in]{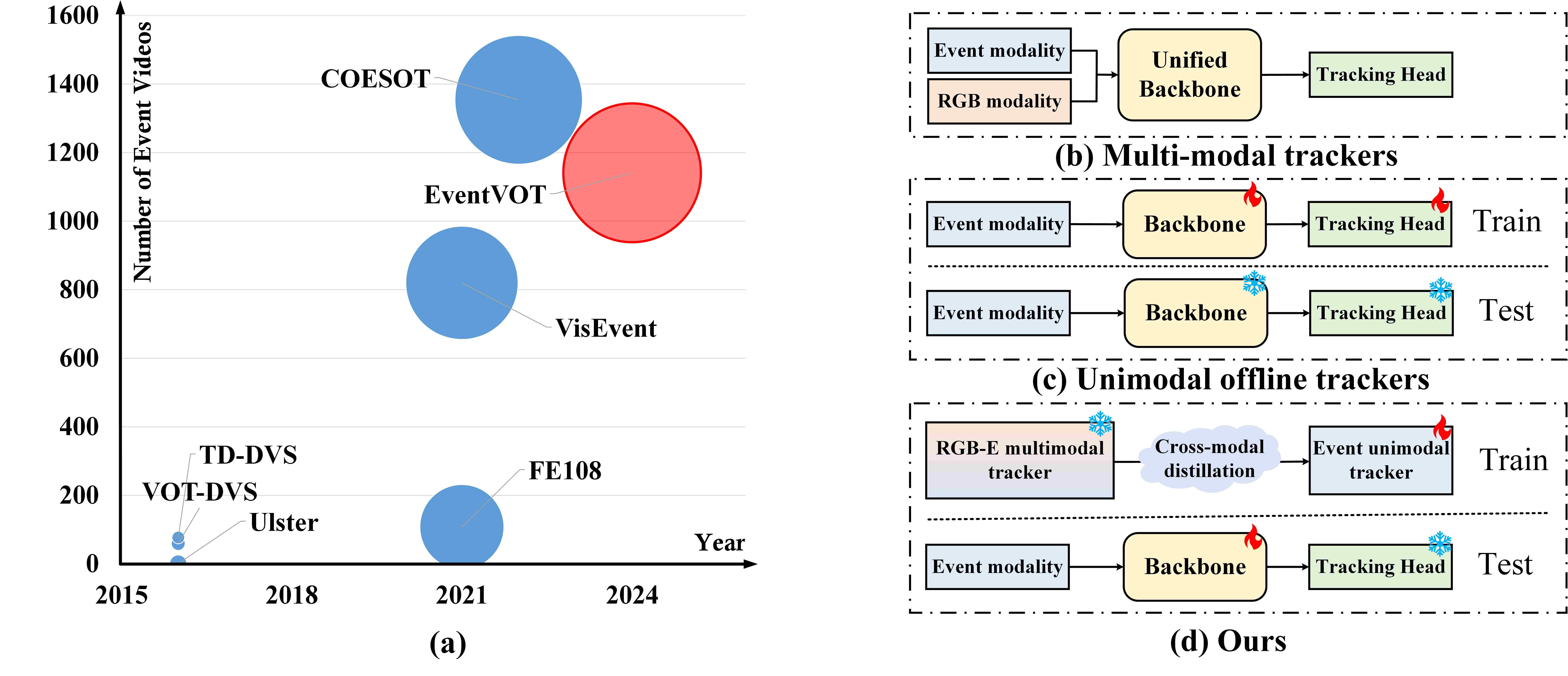}
\caption{ 
(a). Comparison between our newly proposed EventVOT and other event-based tracking datasets; 
(b). RGB-Event based multi-modal tracking framework; 
(c). Pure event-based tracking framework; 
(d). Our newly proposed Test-Time Tuning based event-tracking framework.} 
\label{firstIMG}
\end{figure*}

Recently, some outstanding works have emerged in the field of visual object tracking tasks, leveraging the capabilities of event cameras. For example, Zhang et al. propose  AFNet~\cite{zhang2023AFNet} and CDFI~\cite{zhang2021fe108} to combine the frame and event data via multi-modality alignment and fusion modules. STNet~\cite{zhang2022STN} is proposed to connect the Transformer and spiking neural networks for event-based tracking. Zhu et al.~\cite{zhu2022grapheventTrack} attempt to mine the key events and employ a graph-based network to embed the irregular spatiotemporal information of key events into a high-dimensional feature space for tracking. Messikommer et al.~\cite{messikommer2023data} propose data-driven feature tracking with low-latency event cameras in a gray-scale frame. 
However, these visual trackers still have some unresolved issues: 
\begin{itemize}
    \item \textbf{Multi-modal Tracking Leads to High Computational Costs}: When the target is stationary or moving slowly, event cameras generate sparse signals, blurring the target’s contours and causing tracking failure. Researchers have combined event cameras with RGB cameras to mitigate this issue, but this multi-modal approach significantly increases computational complexity.
    \item \textbf{Offline Training Results in Insufficient Flexibility}: Mainstream tracking frameworks focus on computing the response between the template and search region during the training phase. However, relying solely on this offline mode can't flexibly handle the diverse target objects in actual tracking scenarios, potentially resulting in sub-optimal tracking results. 
    \item \textbf{Lack of High-Resolution Event-based Tracking Datasets}: Current event cameras often produce low-resolution videos (e.g., the DVS346 camera outputs at a resolution of $ 346 \times 260$), which makes it challenging to capture detailed outline information of the target object. However, high-definition event cameras are rarely explored, especially in visual object tracking tasks, as shown in Fig.~\ref{firstIMG} (a).  
\end{itemize}    
Therefore, a thought-provoking question is: \textit{How can we achieve high-performance object tracking using only an event camera?}

Considering the aforementioned challenges, in this work, we propose an object tracking method that uses only the event stream during the inference phase to reduce computational costs, but employs multi-modal data during the training phase to obtain a stronger tracker. Specifically, we propose a novel hierarchical knowledge distillation strategy from multi-modal/multi-view to unimodal based on event cameras for visual object tracking. As illustrated in Fig.~\ref{framework}, we first train a teacher Transformer network using RGB-Event/Event Images-Voxels. The templates and search regions of different modalities/views are cropped by the initial frame and the subsequent frames, then, we adopt a projection layer to transform them into token representations with position encoding. A unified backbone composed of stacked Transformer blocks is used to interact and fuse feature information between different modalities. Finally, we use a center-based tracking head inspired by OSTrack~\cite{ye2022Ostrack} to predict the score maps and bounding boxes in search frames. After the training of the teacher Transformer network, we can obtain a robust but highly complex model. The parameters of the teacher network are frozen, and the student Transformer network is used for effective training. To achieve low latency tracking, only event images are input into the student network and a hierarchical knowledge distillation strategy is performed from the teacher network to enhance the tracking performance. To be specific, the \textit{similarity matrix}, \textit{feature representation}, \textit{response maps}, and \textit{Fourier transformation based knowledge distillation} are simultaneously considered for cross-modality knowledge transfer. 

To further enhance the tracking algorithm’s adaptability to the diverse appearances of target objects in real-world tracking scenarios, we perform \textit{video-level test-time tuning} on the tracker to improve its modeling capabilities for different objects. We assume that the first few frames of a video yield stable tracking results due to relatively minor changes in the scene. Thus, we use their tracking results as ground truth and fine-tune the tracking model. Meanwhile, we propose the consistency constraint to improve the network's generalization ability to prevent the network from falling into local optimization. A comparison between existing (b) RGB-Event based multi-modal tracking, (c) pure event-based tracking, and (d) our newly proposed HDETrack V2 is illustrated in Fig.~\ref{firstIMG}.

To address the lack of high-definition event stream tracking datasets, this paper presents a large-scale, high-definition dataset, termed EventVOT, for this task. Different from existing datasets with limited resolution (e.g., FE240hz~\cite{zhang2021fe108}, VisEvent~\cite{wang2024visevent}, COESOT~\cite{tang2022coesot}, FELT~\cite{wang2024long} are $346 \times 260$) as shown in Fig.~\ref{firstIMG} (a), our videos are collected using the Prophesee camera EVK4–HD which outputs event stream at a resolution of $1280 \times 720$. It contains 1141 videos and covers a wide range of target objects, including pedestrians, vehicles, UAVs, ping pongs, etc. To build a comprehensive benchmark dataset, we provide the tracking results of multiple baseline trackers for future work to compare. We hope our newly proposed EventVOT dataset can open up new possibilities for event tracking.

To sum up, our contributions can be concluded as the following aspects:

$\bullet$ We propose a novel hierarchical cross-modality knowledge distillation strategy for event-based tracking, including similarity-based, feature-based, response-based, and Fourier-based knowledge distillation, termed \textbf{HDETrack V2}. To the best of our knowledge, this is the first work to exploit the knowledge transfer from multi-modal (RGB-Event) / multi-view (Event Image-Voxel) to an unimodal event-based tracker. 

$\bullet$ We propose a test-time tuning strategy that further adapts the network model to specific target objects during testing, achieving improved performance and flexibility in target tracking. 

$\bullet$ We propose the first high-resolution benchmark dataset for event-based tracking termed \textbf{EventVOT}. Extensive experiments are conducted by re-training recent state-of-the-art trackers to build a comprehensive benchmark.  
 
$\bullet$ Extensive experiments on four large-scale benchmark datasets, including FE240hz, VisEvent, FELT, and our proposed EventVOT, fully validated the effectiveness of our proposed tracker.

A preliminary version of this work was published at the international conference, i.e., Computer Vision and Pattern Recognition (CVPR) 2024. 
Compared with the earlier version, i.e., HDETrack~\cite{wang2024event}, we provide novel and valuable ideas, including updates to the methodology framework, detailed explanations of experiments, and further exploration of tracking issues. 
Overall, in our new version, we make the following extensions: 
\textbf{(I). New Knowledge Distillation Strategy:} In addition to the hierarchical knowledge distillation mentioned earlier, we introduce a novel strategy that leverages temporal information by sampling continuous frames and integrating their information using the temporal Fourier transform. 
\textbf{(II). New Test-time Tuning Strategy:} To further improve the generalization of the tracker on the test set, we propose using the Test Time Tuning strategy to enhance the performance of our model. 
\textbf{(III). More Experimental Results and Analysis:} We provide more detailed experiments to validate the effectiveness of our proposed strategy for efficient and effective tracking. Our HDETrack V2 has seen substantial advancements building upon its initial framework, with notable improvements particularly evident in performance on the EventVOT dataset.

\section{Related Work}~\label{sec:Related Work}


\subsection{Event Camera based Tracking}  \label{subsec:eventtracking}
With the development of bio-inspired event cameras, event camera based visual object tracking has gradually become a focus of attention. The early event camera based tracking algorithm ESVM~\cite{huang2018event} proposed an event-guided support vector machine (ESVM) for tracking high-speed moving objects, which solves the tracking problem caused by motion blur and large displacement in low frame rate cameras. Recently, Chen et al.~\cite{chen2019asynchronous} adopt the Adaptive Time Surface with Linear Time Decay (ATSLTD) algorithm for event-to-frame conversion, the spatiotemporal information of asynchronous retinal events is transformed into an ATSLTD frame sequence for efficient object tracking. Zhu et al.~\cite{zhu2022grapheventTrack} propose a new end-to-end learning framework for object tracking based on event cameras, which improves tracking accuracy and speed through key event sampling and graph network embedding. Zhang et al.~\cite{zhang2022STN} propose a novel tracking network STNet based on a spiking neural network and Transformer network, which can effectively extract spatiotemporal information from events and achieve a better tracking performance. Wang et al.~\cite{wang2024event} propose a novel hierarchical knowledge distillation framework, combined with multi-modal/multi-view information, high-speed and low latency visual tracking can be achieved during the testing phase using only event signals.

For event based multi-modal tracking, Zhang et al.~\cite{zhang2021fe108} propose a multi-modal fusion method that combines visual cues from frame and event domains to improve single object tracking performance under degraded conditions, which also enhances the effect through cross-domain attention mechanism and adaptive weighting scheme. 
VisEvent~\cite{wang2024visevent} proposed by Wang et al. transforms the event stream into images and extends a single-modal tracker to a dual-modal version, with a cross-modal converter enabling better fusion of RGB and event data. 
AFNet~\cite{zhang2023AFNet}, proposed by Zhang et al., a framework for high frame rate tracking, event alignment, and fusion network has been proposed, which significantly improves the performance of high frame rate tracking by combining the advantages of traditional frameworks and event cameras.  
Gehrig et al.~\cite{gehrig2020eklt} propose EKLT, an asynchronous photometric feature tracking method that combines the advantages of event cameras and RGB cameras to achieve visual feature tracking at high temporal resolution and improve tracking accuracy. 
ViPT~\cite{zhu2023promptTrack}, which adjusts the pre-trained RGB base model by introducing a small number of trainable parameters to adapt the different multi-modal tracking tasks. 
Different from existing works, we propose to conduct a hierarchical knowledge distillation strategy from multi-modal or multi-view in the training phase and only utilize the event data for efficient and low-latency tracking.

\subsection{Knowledge Distillation}  \label{subsec:kd}
The knowledge distillation strategy has been widely proven to be an effective method of knowledge transfer.
Deng et al.~\cite{deng2021distEvent} propose an image-based knowledge distillation learning framework that improves the performance of event camera models in feature extraction by extracting knowledge from the image domain, achieving significant improvements in the performance of event cameras in target classification and optical flow prediction tasks. 
In visual object tracking, Shen et al.~\cite{shen2021distilledSiamTrack} propose a distillation learning method for learning small, fast, and accurate Siamese network trackers. 
Chen et al.~\cite{chen2022TSKDcorrTrack} propose a lightweight network based on the teacher-student knowledge distillation framework to accelerate visual trackers based on correlation filters while maintaining tracking performance.
Zhuang et al.~\cite{zhuang2021ensembleSiamTrack} propose an ensemble learning method based on Siamese architecture, which effectively improves the accuracy of visual tracking tasks and solves the limitations of knowledge distillation in visual tracking tasks.
Sun et al.~\cite{sun2021USLCMTrack} distill the pre-trained RGB modality onto the TIR modality on unlabeled RGB-TIR datasets, utilize the two branches of the network to process data from different modalities to transfer cross-modal knowledge. 
Wang et al.~\cite{wang2020CFTrack} propose a knowledge distillation framework that combines model compression and transfer. 
Zhao et al.~\cite{zhao2022distSiamTrack} propose a tracking method based on a Siamese network, which optimizes the balance between tracking efficiency and model complexity through distillation, integration, and framework selection, achieving better tracking performance and speed. 
Ge et al.~\cite{ge2019distChannelTrack} propose a novel framework called ``channel distillation", which optimizes the performance of depth trackers by adaptively selecting information feature channels, achieving accurate, fast, and low memory required visual tracking. 
Cui et al.~\cite{cui2024mixformerv2} introduce special prediction tokens and distillation model compression methods in MixFormerV2, which significantly improved efficiency while ensuring accuracy. 
Different from these works, in this paper, we propose a framework for knowledge distillation from multi-modal or multi-view to unimodal, which achieves efficient visual object tracking through hierarchical knowledge distillation.

\subsection{Test-Time Tuning} \label{subsec:ttt} 
The idea of Test Time Tuning is to use self-supervised learning to continue training the model on test sets. 
TTT~\cite{sun2020test} proposed by Yu et al., introduces a Test Time Tuning method that improves the performance of prediction models under data distribution shift through self-supervised learning.
In the field of computer vision, Test Time Tuning has also been proven to be an effective strategy in many tasks~\cite{jain2011online, mullapudi2019online, nitzan2022mystyle, shocher2018zero, tonioni2019learning, tonioni2019real, zhang2020online, zhong2017recovery, zhong2018open}. 
Yossi et al.~\cite{gandelsman2022test} propose TTT-MAE, which by studying and comparing three different training strategies during the training phase, ViT-probing is ultimately selected as the appropriate training method. In the testing phase, the image reconstruction task was used to further optimize the image encoder on the test set, achieving significant performance improvement.
Wang et al.~\cite{wang2023test} introduce the TTT strategy into video modeling and propose online TTT. 
Mirza et al.~\cite{mirza2023actmad} propose a new Test Time Tuning method called ActMAD, which aligns distributions through matching activations to address distribution shift issues and achieve performance improvements across multiple tasks and architectures.
Liu et al.~\cite{liu2024depth} propose a Test Time Tuning method for zero-shot video object segmentation, which significantly improves segmentation performance by training the model to predict consistent depth during testing.
In this paper, we introduce the Test Time Tuning strategy in our tracker, by optimizing the model through consistency constraints during the inference process to improve its generalization ability on the test set.

\begin{figure*}[!htp]
\center
\includegraphics[width=7in]{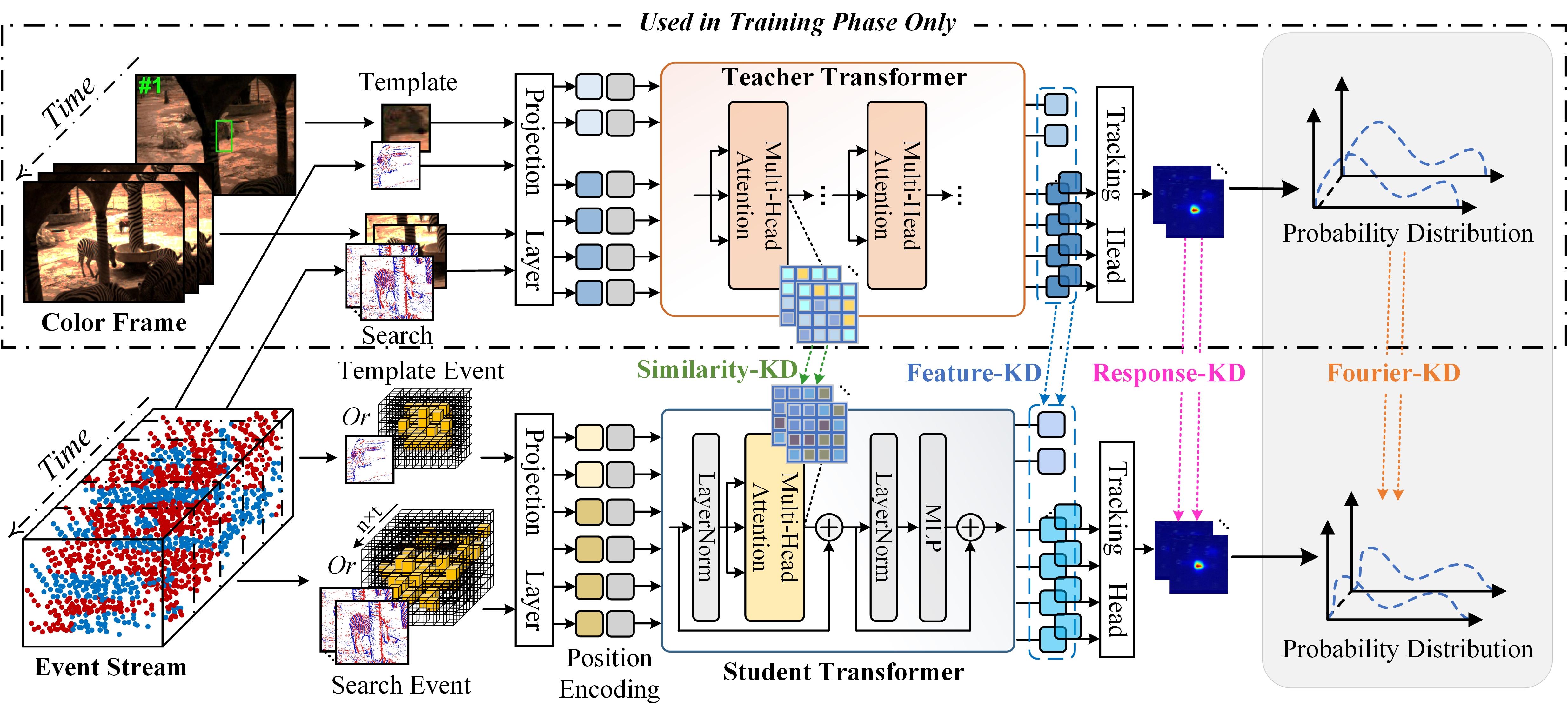}
\caption{\textbf{An overview of our proposed \textbf{H}ierarchical Knowledge \textbf{D}istillation Framework for \textbf{E}vent Stream based Tracking, termed HDETrack V2.} It contains the teacher and student Transformer networks which take multi-modal/multi-view and event data only as the input respectively. Both networks share an identical architecture, i.e., tracking using a unified Transformer backbone network similar to CEUTrack~\cite{tang2022coesot} and OSTrack~\cite{ye2022Ostrack}. Specifically, we extract the template and search patches from both RGB and event inputs, generating feature embeddings through a projection layer. These embeddings are then passed through a stack of Transformer layers that form the teacher network. The output of this network feeds into the tracking head for target object localization. Meanwhile, the student network is designed for efficient tracking using event stream only. 
It is trained with tracking loss functions and benefits from knowledge distillation from the teacher Transformer network. 
Our tracker achieves a better tradeoff between accuracy and model complexity.} 
\label{framework}
\end{figure*}

\section{METHODOLOGY}~\label{sec:Method}

In this section, we will first give an overview of our proposed HDETrack V2. Then, we will introduce the details from the input representation, network architecture, hierarchical knowledge distillation strategy, and test time tuning in the inference phase.

\subsection{Overview} 
As shown in Fig.~\ref{framework}, our proposed HDETrack V2 follows the teacher-student framework, where the teacher network is the RGB-Event tracking subnetwork and the student network is the Event-based tracking network. The teacher network takes the RGB frames and Event streams as the input and extracts the multi-modal features using a unified Transformer network. The student network takes the Event stream as the input only which can achieve efficient tracking in the inference phase. More importantly, we transform the knowledge from the teacher network to the student via a hierarchical knowledge distillation strategy, i.e., similarity matrix-based, feature-based, response-based, and temporal Fourier-based distillation. Note that, the teacher network is only used in the training phase. In the tracking phase, we also propose to adapt the neural network to the diverse target objects using a test time tuning scheme, as shown in Fig.~\ref{framework_TTT}. More details will be introduced in the following sub-sections respectively.


\subsection{Input Representation}~\label{subsec:input}
In this work, we denote the RGB frames as $\mathcal{I} = \{I_1, I_2, ..., I_N\}$, where $I_i \in \mathbb{R}^{1280 \times 720}$ represents each video frame, $i \in [1, N]$, where $N$ is the number of video frames. We treat event stream as $\mathcal{E} = \{e_1, e_2, ..., e_M\}$, with $e_j$ denoting each asynchronously launched event point, $j \in [1, M]$, and $M$ is the number of event points in the current sample. 
For the video frames $\mathcal{I}$, we apply standard Siamese tracking methods to extract the template $T_I \in \mathbb{R}^{128 \times 128}$ and search $S_I \in \mathbb{R}^{256 \times 256}$ as input. 
For the event stream $\mathcal{E} \in \mathbb{R}^{1280 \times 720}$, we stack/split them into event images/voxels which can fuse more conveniently with existing RGB modality. Specifically, the event images are obtained by aligning with the exposure time of the RGB modality. Event voxels are obtained by splitting the event stream along with the spatial (width $W$ and height $H$) and temporal dimensions ($T_i$). The scale of each voxel grid is denoted as ($a, b, c$), thus, we can get $\frac{W}{a} \times \frac{H}{b} \times \frac{T_i}{c}$ voxel grids. Similarly, we can obtain the template and search regions of event data, i.e., $T_E \in \mathbb{R}^{128 \times 128}$ and $S_E \in \mathbb{R}^{256 \times 256}$.

\subsection{Network Architecture}~\label{subsec:pipeline}

In this paper, our proposed hierarchical knowledge distillation framework contains the Multi-modal/Multi-view Teacher Transformer and Unimodal Student Transformer network for event-based tracking.

\subsubsection{Multi-modal/Multi-view Teacher Tracker} 
For the teacher network, we take both the RGB frames and event streams as the input to obtain more comprehensive information representations. If no RGB frames are available (e.g., the EventVOT dataset), we input multi-view event representations such as the event images and event voxels. Specifically, using the RGB modality as an example (the same applies to the event modality), we randomly sample a template frame and a search frame from a video to form a pair of samples. These frames are partitioned into image patches according to the specified patch size ($16 \times 16$) and transformed into token embedding ($64 \times 768$ and $256 \times 768$ for template and search tokens) via a projection layer (convolutional layer with kernel size $16 \times 16$). Subsequently, we add position encoding to label the position of each token in the entire patch sequence. Next, we concatenate the template and search tokens together along the dimension of the number of tokens, and input them into the unified Transformer backbone network by following~\cite{ye2022Ostrack, tang2022coesot} for interaction and fusion of features. We further decompose the search features from the entire output and input the search features to the tracking head for target object localization and tracking.

\subsubsection{Unimodal Student Tracker} 
After the first stage of training, we can obtain a robust teacher Transformer network trained on multi-modal/multi-view data. To enable efficient visual tracking, we employ knowledge distillation to transfer essential knowledge from the teacher network to the student network, allowing us to use only a lightweight student network for effective inference independently during the testing phase. 
As illustrated in Fig.~\ref{framework}, only event data is fed into the student Transformer for tracking. 
Our student network's processing follows a similar structure to the teacher network, but only event data is fed into the student Transformer network, thus achieving a significant speed advantage. Furthermore, thanks to the carefully designed hierarchical knowledge distillation, our student network can learn more useful knowledge through the guidance of the teacher network beyond standard ground truth-based supervised learning, thereby improving the accuracy of tracking.

\subsubsection{Hierarchical Knowledge Distillation}  
In addition to the training losses used in OSTrack~\cite{ye2022Ostrack}, i.e., focal loss $\mathcal{L}_{focal}$, $L_1$ loss $\mathcal{L}_{L1}$, and GIoU loss $\mathcal{L}_{GIoU}$), we also introduce four additional Knowledge Distillation (KD) losses to further guide the optimization of the student network from our pre-trained teacher network. Specifically, it includes three modules for spatial distillation: similarity matrix-based distillation, feature-based distillation, and response map-based knowledge distillation, as well as temporal Fourier transform distillation.

\begin{figure*}[!htp]
\center
\includegraphics[width=6in]{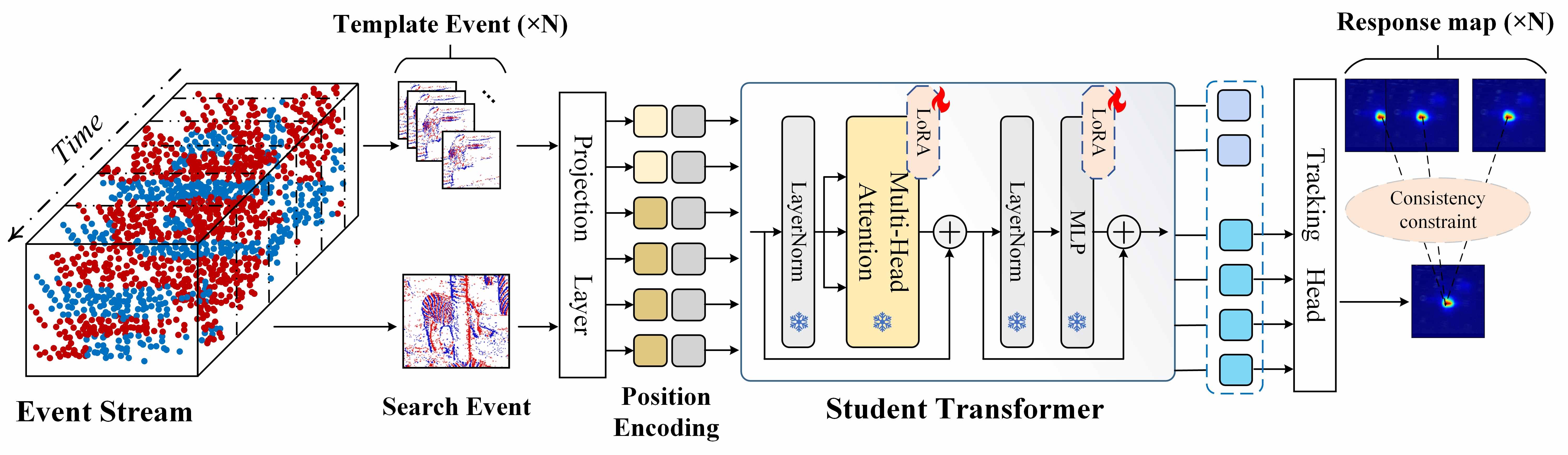}
\caption{\textbf{The Test Time Tuning (TTT) strategy employed during the inference phase.} The template frames are augmented based on the sparsity of the initial event streams and then fused within the search region to yield a variety of response maps. We further enhance the tracker's efficacy through the integration of LoRA specifically tailored for the testing phase. It is worth noting that, to maintain alignment between the TTT stage and the foundational training stage, we achieve it by using the tracking results of the initial few frames in a video as the pseudo labels for self-supervised learning. For more detailed implementation, please refer to section~\ref{subsec:TTTT}.}
\label{framework_TTT}
\end{figure*}

\noindent $\bullet$ \textbf{\textit{Similarity Matrix based KD: }} 
It is computed by the multi-head self-attention layers incorporating abundant long-range and cross-modal relation information. In this work, we exploit the knowledge transfer from the similarity matrix learned by the teacher Transformer to the student Transformer. The loss function can be written as:
\begin{equation}
 \label{similarityMatrixDistill} 
 \mathcal{L}_{simKD} = \sum (S_s^j - S_t^i)^2,  
\end{equation}
where $S_t^i \in \mathbb{R}^{640 \times 640}$ denotes the similarity matrix of the $i^{th}$ teacher Transformer layer, and $S_s^j \in \mathbb{R}^{320 \times 320}$ denotes the similarity matrix of the $j^{th}$ student Transformer. 
To align the dimensions of the two similarity matrices, we perform a simple repeat approach on $S_s^j$ and use it to calculate the distance between the two matrices.

\noindent $\bullet$ \textbf{\textit{Feature based KD: }} 
This strategy has demonstrated its superiority in numerous works. Therefore, we further perform feature-based distillation between the teacher and student networks. Specifically, we utilize the \textit{Mean Square Error} (MSE) loss to transfer the knowledge. The loss function can be represented as:
\begin{equation}
\label{featDistill} 
\mathcal{L}_{featKD} = \frac{1}{N} \sum_{i=1}^{N} (F_t - F_s)^2, 
\end{equation}
where $F_t \in \mathbb{R}^{B \times 640 \times 768}$ and $F_s \in \mathbb{R}^{B \times 320 \times 768}$ are the output features of the teacher network and student network, respectively. Similar to the above, we also use repeat operations to achieve alignment between two features, which has been proven to be effective.

\noindent $\bullet$ \textbf{\textit{Response Map based KD: }} 
The response map serves as an important reference for tracking networks. We introduce response-based distillation to enable our student network to mimic the teacher network and produce more accurate output results. In this paper, the weighted focal loss function~\cite{law2018cornernet} is adopted to achieve this target. We denote the ground truth target center and the corresponding low-resolution equivalent as $\hat{p}$ and $\bar{p} = [ \bar{p}_x, \bar{p}_y ]$, respectively. The Gaussian kernel is used to generate the ground truth heatmap $\hat{\bf{P}}_{xy} = exp (- \frac{(x-\bar{p}_x)^2 + (y-\bar{p}_y)^2}{2\delta^2_p})$, where $\delta$ denotes the object size-adaptive standard. The Gaussian Weighted Focal (GWF) loss function can be formulated as:
\begin{equation}
\label{GaussianWFocalLoss} 
\small 
\mathcal{L}_{GWF} = - \sum_{xy} 
\begin{cases}
(1-\textbf{P}_{xy})^\alpha log(\textbf{P}_{xy}),   & if~\hat{\textbf{P}}_{xy} = 1   \\
(1-\hat{\textbf{P}}_{xy})^\beta(\textbf{P}_{xy})^\alpha log(1-\textbf{P}_{xy}),   &otherwise 
\end{cases}
\end{equation}
where $\alpha$ and $\beta$ are two hyper-parameters and are set to 2 and 4 respectively in our experiments, as suggested in OSTrack~\cite{ye2022Ostrack}. In our implementation, we normalize the response maps of both the teacher and student networks by dividing them by a temperature coefficient $\tau$ (empirically set to 2), followed by inputting them into the focal loss for response distillation, i.e., $\mathcal{L}_{resKD} = \mathcal{L}_{GWF}(R_s/\tau, R_t/\tau )$.

\noindent $\bullet$ \textbf{\textit{Temporal Fourier Transform based KD: }}   
Drawing inspiration from the extant studies~\cite{xie2024autoregressive, cao2022tctrack}, we acknowledge the critical role of encapsulating the temporal dynamics between consecutive video frames for the advancement of video tracking. This work incorporates the temporal relationships between frames into the knowledge distillation. To establish the temporal relationship between frames from the perspective of distillation, we propose a novel knowledge distillation strategy based on the temporal Fourier transform. Specifically, our breakthrough point is to sample $one$ reference frame and $n$ consecutive frames from each video as the template and search regions to train our student network during the knowledge distillation stage. Thus, we can obtain $\{T_0, S_1, S_2, ..., S_n\}$ for a video sequence. However, it does not bring ideal results due to the lack of interactive fusion between frames.

Consequently, we propose establishing the interaction between frames through the Fourier transform. By inputting $n$ frames of search regions, we can obtain $n$ score maps corresponding to each frame, which reflect the confidence score of the tracking results on these frames, represented with $Map_i, i \in [s, t]$. Subsequently, we perform the $softmax$ function on these score maps to obtain the probability distribution of each tracking result. The Fourier transform is then applied to these probability distributions, converting frequency information into corresponding temporal information to establish the Fourier temporal relationship between input frames. Thus, we obtain $C_i (i \in [s, t])$ with temporal correlation from both student and teacher networks. Note that we extract the corresponding real parts from the Fourier transform results to facilitate knowledge transfer from teachers to students. The formula is as follows：
\begin{align}
\label{TFTLoss} 
\mathcal{L}_{tftKD} &= \mathcal{L}_{GWF}(C_s, C_t), \\
C_i &= f(\textit{softmax}(Map_i)), \quad i \in [s, t], \\
f(x) &= x_{m,n} = \frac{1}{M N} \sum_{k=0}^{M-1} \sum_{l=0}^{N-1} X_{k,l} e^{i \frac{2 \pi}{M} \left( k m + l n \right)}, 
\end{align}
where $M$ and $N$ are the spatial size of the image used to represent the number of sampling points in the horizontal direction (width) and vertical direction (height). $k$ and $l$ are frequency index variables, which represent the frequency positions in the horizontal and vertical directions, their value ranges from $0$ to $M-1$ and from $0$ to $N-1$, respectively. $m$ and $n$ represent spatial position index variables. 

Finally, the total loss function during training with knowledge distillation can be expressed as:
\begin{align}
    \label{lossFunction} 
    & \mathcal{L}_{total} = \lambda_1 \mathcal{L}_{focal} + \lambda_2 \mathcal{L}_{1} + \lambda_3 \mathcal{L}_{GIoU} + \\ 
    &~~~~~~~~~~~~~~~~ \eta_1 \mathcal{L}_{simKD} + \eta_2 \mathcal{L}_{featKD} + \eta_3 \mathcal{L}_{resKD}   \nonumber + \eta_4 \mathcal{L}_{tftKD}.
\end{align}
Here, $\lambda_1$, $\lambda_2$, and $\lambda_3$ are the weight coefficients of the main losses like CEUTrack~\cite{tang2022coesot}. $\eta_1$, $\eta_2$, $\eta_3$, $\eta_4$ are hyper-parameters in knowledge distillation strategies.

\subsection{Test Time Tuning in Tracking\label{subsec:TTTT}}
To alleviate the distribution gap between the training set and the test set, which impacts the model's generalization ability, we adopted the Test Time Tuning strategy, as shown in Fig.~\ref{framework_TTT}. Specifically, in order to reduce the cost of training during the testing phase, we freeze the trained network and integrate LoRA~\cite{hu2022lora}, which has proven effective for fine-tuning models, especially large language models, to further optimize the model on the test set. Given the absence of labels during testing, we can only employ self-supervised or unsupervised learning. 

In this work, based on empirical insights, we observe that the target object's movement is relatively small and interference is minimal in the initial stages of tracking. To this end, we use the tracking results from these early frames as pseudo-labels to facilitate self-supervised training. We first perform normal tracking on the initial few frames of a video, recording the results of these frames as 'ground truth'. Subsequently, we leverage these frames and their corresponding 'ground truth' to further train our model. In addition, to prevent the model from converging to local optima, we introduce regularization terms to reduce over-fitting. For regularization, we augment the initial frame of each test video, taking into account the characteristics of event data and the significance of the template frame in tracking tasks. Specifically, we redivide the initial frame into $n$ template frames according to varying levels of event sparsity, and we calculate $n$ corresponding response maps based on the $i$-th search. Finally, we build the consistency constraint loss among these $n$ response maps. To address the challenge of the target object moving too quickly or jumping out of the field of view during the testing phase, we incorporate a simple yet effective \textit{Adaptive Search Region strategy} inspired by~\cite{tang2023learning, wang2021TANet}. This approach expands the ratio of crop size of the search region by $\theta$ times when the IoU values between consecutive tracking results fall below a specified threshold $\tau$. By implementing a series of tuning strategies during testing, we can further enhance the performance of our model on an existing basis.

\section{EventVOT Benchmark Dataset}\label{sec:EventVOT Dataset}
In this paper, we propose a large-scale high-definition event-based visual tracking dataset. Here, we introduce the protocols for data collection and annotation, statistical analysis, and benchmarked visual trackers in the following subsections, respectively. 

\begin{figure*}[!htp]
\center
\includegraphics[width=\linewidth]{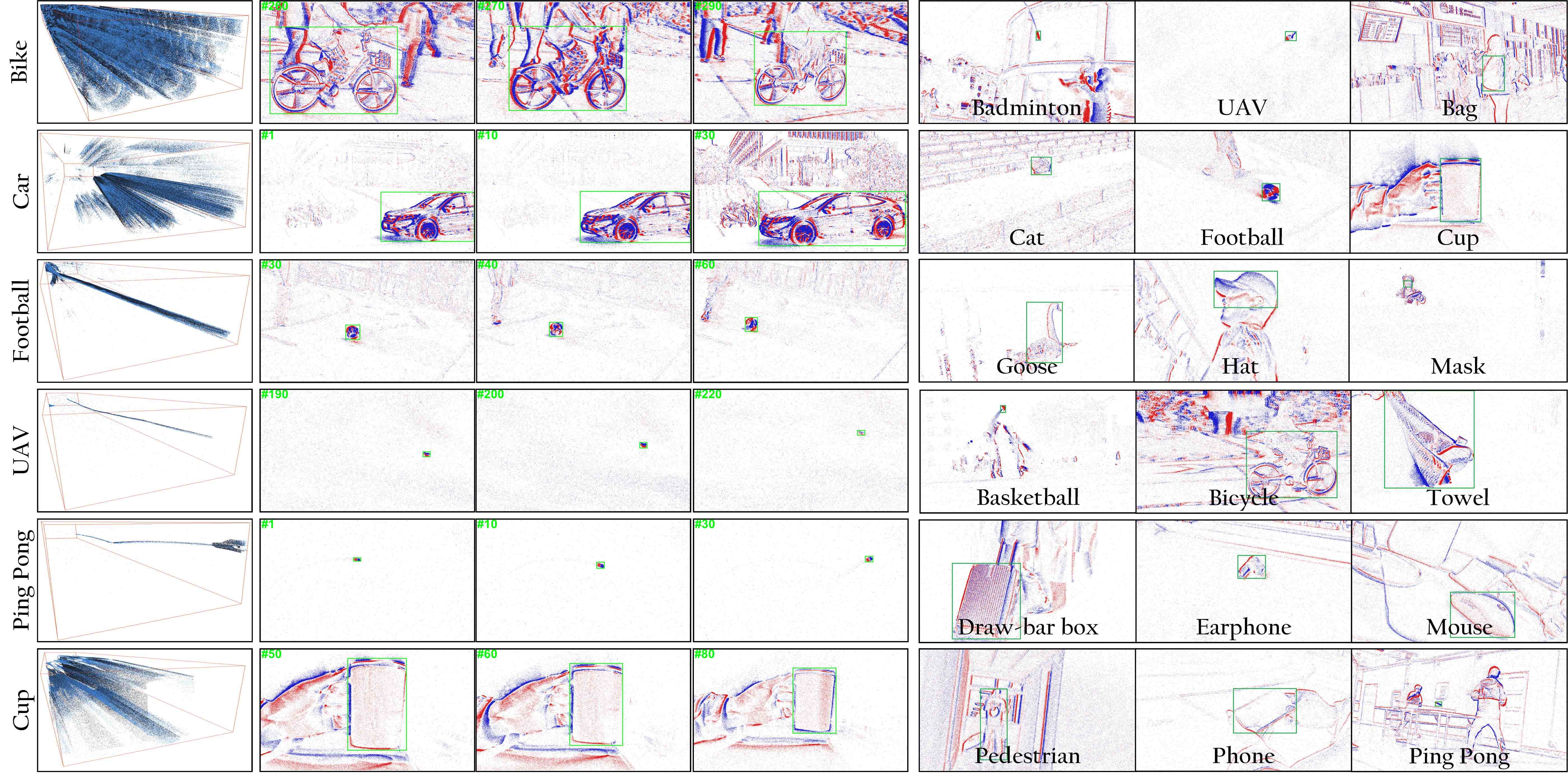}
\caption{Representative samples of our proposed EventVOT dataset. The $1^{th}$ column is the 3D event point stream and the $2^{th}$-$4^{th}$ columns are sampled event images. $5^{th}$-$7^{th}$ columns are more samples of our EventVOT dataset.}   
\label{eventvotExamples}
\end{figure*}

\begin{figure*}[!htp]
\center
\includegraphics[width=\linewidth]{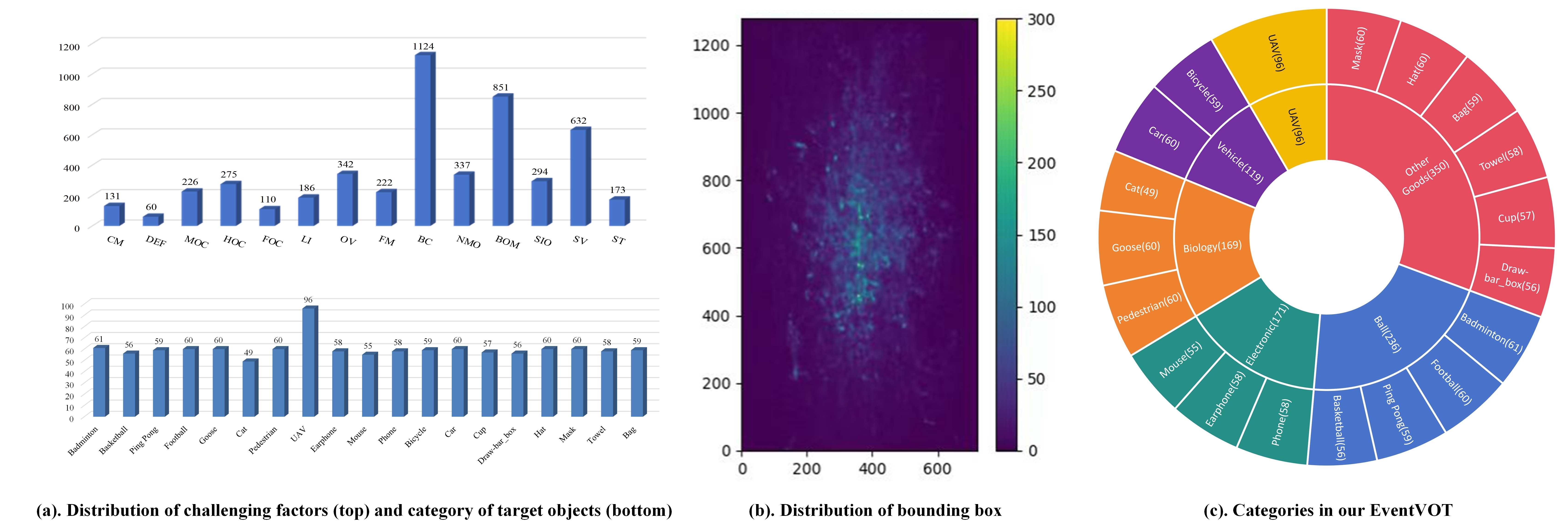}
\caption{Distribution visualization of challenging factors, category of the target object, and bounding box.} 
\label{datasetInfo}
\end{figure*}

\begin{table*}
\center
\footnotesize
\caption{Event camera based datasets for visual tracking. \# denotes the number of corresponding items.} 
\label{benchmarkList}
\resizebox{\textwidth}{!}{ 
\begin{tabular}{l|cccccccccccccccc}
\hline \toprule [0.5 pt]
\textbf{Datasets}    &\textbf{Year}	&\textbf{\#Videos}  &\textbf{\#Frames} &\textbf{\#Class}    &\textbf{\#Att} &\textbf{\#Resolution} &\textbf{Aim}   &\textbf{Absent}  &\textbf{Frame} &\textbf{Reality}  &\textbf{Public}  \\ 
\hline
\textbf{VOT-DVS}     &2016    &60           &-      &-    	 &-    &$240 \times 180$  &Eval  &\xmark   &\xmark     &\xmark     &\cmark     \\
\textbf{TD-DVS}        &2016     &77          &-      &-    	 &-  &$240 \times 180$  &Eval  &\xmark    &\xmark     &\xmark   &\cmark     \\
\textbf{Ulster}  &2016      &1     &9,000  		   &-    	 &-    &$240 \times 180$ 
 &Eval   &\xmark  &\xmark     &\cmark     &\xmark    		\\
\textbf{EED} &2018     &7     &234   &  -  	 &-   &$240 \times 180$   &Eval  &\xmark   &\xmark      &\cmark     &\cmark     \\
\hline 
\textbf{FE108}	&2021		&$108$     &208,672  	& 21  & 4     &$346 \times 260$     &Train \& Eval   &\xmark     &\xmark     &\cmark       &\cmark    \\
\textbf{VisEvent}  	&2021     &$820$       &371,127      	&  -  	&\textbf{17}     &$346 \times 260$  &Train \& Eval      &\cmark &\cmark          &\cmark           &\cmark     \\
\textbf{COESOT}  	&2022     &\textbf{1354}     &478,721       &\textbf{90}    	&\textbf{17}   &$346 \times 260$    & Train \& Eval   &\cmark &\cmark  &\cmark   &\cmark     \\
\textbf{FELT}   &2024  &742  &\textbf{1,594,474}  & 45 & 14 & $346 \times 260$ & Train \& Eval &\cmark &\cmark  &\cmark   &\cmark     \\
\hline
\textbf{EventVOT}    &2024     &1141     &569,359       &19    	&14   &\textbf{1280} $\times$ \textbf{720}    & Train \& Eval  &\cmark &\xmark  &\cmark   &\cmark     \\ 
\hline \toprule [0.5 pt]
\end{tabular}
}
\end{table*}

\subsection{Criteria for Collection and Annotation\label{subsec:Collection and Annotation}} 
To construct a dataset with a diverse range of target categories, as shown in Fig.~\ref{eventvotExamples}, capable of reflecting the distinct features and advantages of event tracking, this paper primarily considers the following aspects during data collection. 
\emph{1). Diversity of target categories:} Many common and meaningful target objects are considered, including UAVs, pedestrians, vehicles, ball sports, etc. 
\emph{2). Diversity of data collection environments:} The videos in our dataset are recorded in day and night time, and involved venue information includes playgrounds, indoor sports arenas, main streets and roads, cafeteria, dormitory, etc. 
\emph{3). Recorded specifically for event camera characteristics:} Different motion speeds, such as high-speed, low-speed, momentary stillness, and varying light intensity, etc. 14 challenging factors are reflected by our EventVOT dataset. 
\emph{4). High-definition, wide-field event signals:} The videos are collected using a Prophesee EVK4–HD event camera, which outputs event stream with $1280 \times 720$. This high-definition event camera excels in supporting pure event-based object tracking, thereby avoiding the influences of the RGB cameras and showcasing its features and advantages in various aspects such as high-speed, low-light, low-latency, and low-power consumption.   
\emph{5). Data annotation quality:} All data samples are annotated by a professional data annotation company and has undergone multiple rounds of quality checks and iterations to ensure the accuracy of the annotations. For each event stream, we first stack into a fixed number (499 in our case) of event images for annotation. 
\emph{6). Data size:} Collect a sufficiently large dataset to train and evaluate robust event-based trackers. 
A comparison between the newly proposed dataset and existing tracking datasets is summarized in Table~\ref{benchmarkList}.

\subsection{Statistical Analysis\label{subsec:Statistical Analysis}} 
In the EventVOT dataset, we have defined 14 challenging factors, involving 19 classes of target objects. The number of videos corresponding to these attributes and categories is visualized in Fig.~\ref{datasetInfo} (a, c). We can find that BC, BOM, and SV are top-3 major challenges which demonstrates that our dataset is relatively challenging. The balance between different categories is also well-maintained, with the number of samples roughly distributed between 50 to 60. Among them, UAVs (Unmanned Aerial Vehicles) are a special category of targets, with a total count of 96. The distribution of the center points of the annotated bounding boxes is visualized in Fig.~\ref{datasetInfo} (b). 
Our EventVOT dataset is split into training/validation/testing subsets which contain 841, 18, and 282 videos, respectively.

\subsection{Benchmarked Trackers\label{subsec:Benchmark}} 
To build a comprehensive benchmark on the EventVOT dataset for event-based visual tracking, we consider more than 20 visual trackers: 
\textbf{1). Siamese or Discriminate trackers:} DiMP50~\cite{goutam2019Dimp}, PrDiMP~\cite{martin2020PrDimp}, 
KYS~\cite{bhat2022SKys}, ATOM~\cite{martin2019Atom}, 
\textbf{2). Transformer trackers:} OSTrack~\cite{ye2022Ostrack}, TransT~\cite{chen2021transt}, SimTrack~\cite{chen2022SimTrack}, AiATrack~\cite{gao2022AIa}, STARK~\cite{yan2021Stark}, ToMP50~\cite{mayer2022Tomp}, MixFormer~\cite{cui2022Mixformer}, TrDiMP~\cite{wang2021TrDiMP}, ROMTrack~\cite{cai2023robust}, CiteTracker~\cite{li2023citetracker}, ARTrack~\cite{wei2023autoregressive}, AQATrack~\cite{xie2024autoregressive}, ODTrack~\cite{zheng2024odtrack}, EVPTrack~\cite{shi2024explicit}, ARTrackV2~\cite{bai2024artrackv2}, LoRAT~\cite{lin2024tracking}. 
Note that, we re-train these trackers using their default settings on the training dataset for a fair comparison, instead of directly testing on the testing subset. The benchmark can be found in Table~\ref{EventVOT_auc}. We believe that these retrained tracking algorithms can play a crucial role in future comparisons of their performance.

\begin{table}
\center
\small  
\caption{Description of 14 attributes in our EventVOT dataset.} 
\label{AttributeList}
\begin{tabular}{l|lcccccccccccccc}
\hline \toprule [0.5 pt]
\textbf{Attributes}    &\textbf{Description}  \\ 
\hline
\textbf{01. CM}   	    	&Abrupt motion of the camera \\	
\textbf{02. MOC}   	    &Mildly occluded \\	
\textbf{03. HOC}   	    &Heavily occluded \\
\textbf{04. FOC}   	    &Fully occluded \\
\textbf{05. DEF}   	    &The target is deformable \\	
\textbf{06. LI}   	    	&Low illumination \\ 
\textbf{07. OV}   	    	&The target completely out of view \\ 
\textbf{08. SV}   	    	&Scale variation  \\
\textbf{09. BC}   	    	&Background clutter  \\
\textbf{10. FM}   	    	&Fast motion  \\
\textbf{11. NMO}   	    &No motion  \\
\textbf{12. BOM}         &Influence of background object motion \\ 
\textbf{13. SIO}         &Similar interferential object \\ 
\textbf{14. ST}          &Small target \\ 
\hline \toprule [0.5 pt]
\end{tabular}
\end{table}


\section{Experiments}~\label{sec:Experiments}

\subsection{Dataset and Evaluation Metric\label{Dataset}} 
In addition to our newly proposed \textbf{EventVOT} dataset, we also compare our tracker with other SOTA visual trackers on existing event-based tracking datasets, including \textbf{FE240hz}~\cite{zhang2021fe108}, \textbf{VisEvent}~\cite{wang2024visevent}, and \textbf{FELT}~\cite{wang2024long} dataset. A brief introduction to these event-based tracking datasets is given below. 

\noindent $\bullet$ \textbf{FE240hz dataset}: It is collected using a gray-scale DVS346 event camera which contains 71 training videos and 25 testing videos. More than 1132K annotations on more than 143K images and corresponding events are provided. It considers different degraded conditions for tracking, such as motion blur and high dynamic range. 

\noindent $\bullet$ \textbf{VisEvent dataset}: It is the first large-scale frame-event tracking dataset recorded using a color DVS346 event camera. A total of 820 videos are collected in both indoor and outdoor scenarios. Specifically, the authors split these videos into a training subset and a testing subset which contain 500 and 320 videos, respectively.


\noindent $\bullet$ \textbf{FELT dataset}: It is the first long-term tracking dataset, with each video shot lasting about 3 minutes and containing a total of 1,594,474 video frames. 14 challenge attributes are considered in this dataset. The training and testing subset contains 520 and 222 videos, respectively.

For the evaluation metrics, we use widely adopted measures such as Precision Rate (PR), Normalized Precision Rate (NPR), and Success Rate (SR). Efficiency is also a key consideration for practical trackers, in this work, we use Frames Per Second (FPS) to assess the speed of each tracker.

\subsection{Implementation Details\label{Implementation Details}} 
The training of our tracker can be divided into two stages. We first pre-train the teacher Transformer with multi-modal/multi-view inputs for 50 epochs. The learning rate is 0.0001, weight decay is 0.0001, and batch size is 32. Subsequently, we adopt a hierarchical knowledge distillation strategy for training the student Transformer network. The learning rate, weight decay, and batch size are set to 0.0004, 0.0001, and 32, respectively. 
The AdamW optimizer~\cite{loshchilov2018adamw} is selected for the training of our tracking network. In the testing phase, we adopt the TTT strategy to enhance the model's generalization on the test set. The Stochastic Gradient Descent (SGD) is selected as the optimizer. The learning rate, weight decay, and epoch are set to 0.01, 0.1, and 5, respectively. 
Our code is implemented using Python based on PyTorch~\cite{paszke2019pytorch} framework and the experiments are conducted on a server with CPU Intel(R) Xeon(R) Gold 5318Y CPU @2.10GHz and GPU RTX3090. 
More details can be found in our source code on GitHub.

\begin{figure*}[!htp]
\center
\includegraphics[width=\linewidth]{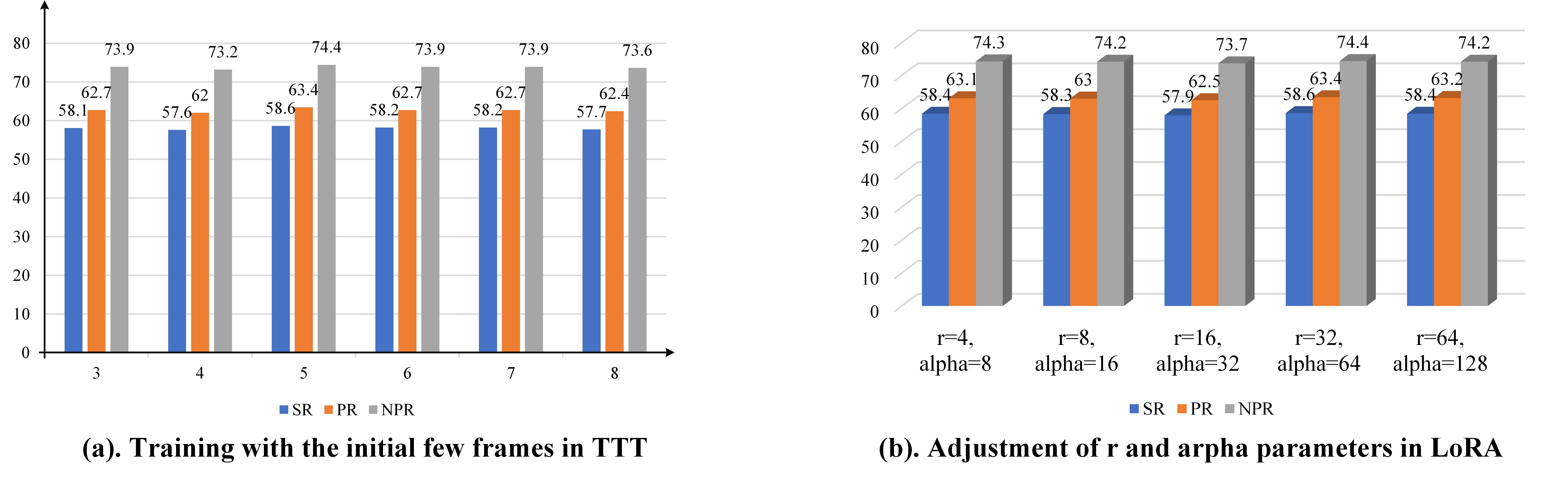}
\caption{(a). Training with the initial few frames in TTT; (b) Adjustment of $r$ and $alpha$ parameters in LoRA.} 
\label{ablation1}
\end{figure*}

\begin{figure*}
\center
\includegraphics[width=7in]{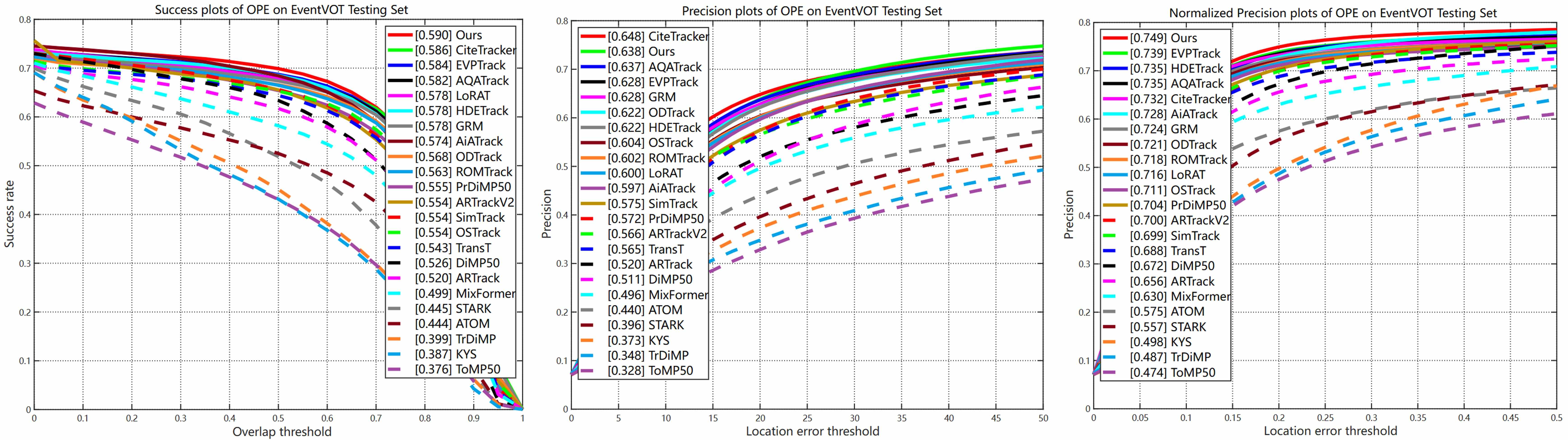}
\caption{Visualization of tracking results of our proposed EventVOT dataset.}  
\label{PRSRNPRfig}
\end{figure*}

\subsection{Comparison on Public Benchmarks\label{Comparison on Public Benchmarks}} 

\noindent $\bullet$ \textbf{Results on EventVOT Dataset.\label{EventVOT}}  
As shown in Table~\ref{EventVOT_auc} and Fig.~\ref{PRSRNPRfig}, we further incorporate several new SOTA tracking methods for comparison, building on our previous benchmark. To ensure a fair comparison, we re-train and evaluate these methods on the EventVOT dataset. Firstly, we can see that our newly designed HDETrack V2 has significantly surpassed our baseline method OSTrack, achieving $59$, $63.8$, and $74.9$ on the SR, PR, and NPR, respectively. Furthermore, compared to HDETrack, the upgraded HDETrack V2 has also been further improved in various metrics, with $+1.2, +1.6$, and $+1.4$ improvements on SR, PR, and NPR, which fully validates the effectiveness of our proposed method for event-based tracking. At the same time, our method also outperforms all SOTA trackers, including the Siamese trackers and Transformer trackers (e.g., ODTrack, EVPTrack, ARTrackV2, AQATrack) on NPR, while the SR ranks only behind the optimal AQATrack and PR also ranking second. These experimental results highlight the effectiveness of our series of improvements to the HDETrack, including the incorporation of Temporal Fourier Transform, Test-Time Tuning, and Adaptive Search Region strategies. Similar conclusions can also be drawn from the experimental results on FE240hz (Table~\ref{FE240table}), VisEvent (Table~\ref{Viseventtable}), and FELT (Table~\ref{FELT_benchmark_results}).

\begin{table}
\center
\small   
\caption{Overall Tracking Performance on EventVOT Dataset. } 
\label{EventVOT_auc}
\resizebox{\columnwidth}{!}{ 
\begin{tabular}{l|l|l|lll|ll}
\hline \toprule [0.5 pt]
\textbf{No.} & \textbf{Trackers} & \textbf{Source}   & \textbf{SR}  &\textbf{PR}   &\textbf{NPR}  &\textbf{Params}  &\textbf{FPS}\\
\hline
01    &  \textbf{ DiMP50~\cite{goutam2019Dimp} }  &  ICCV19       &\ 52.6   &\ 51.1   &\ 67.2   &\ 26.1  &\ 43  \\
02    &  \textbf{ ATOM~\cite{martin2019Atom}   }   & CVPR19     &\ 44.4   &\ 44.0   &\ 57.5   &\ 8.4   &\ 30  \\
03    &  \textbf{ PrDiMP~\cite{martin2020PrDimp}}  &  CVPR20       &\ 55.5   &\ 57.2   &\ 70.4   &\ 26.1   &\ 30  \\
04    &  \textbf{ KYS~\cite{bhat2022SKys}   }   &   ECCV20         &\ 38.7   &\ 37.3   &\ 49.8   &\ --   &\ 20  \\ 
05    &  \textbf{ TrDiMP~\cite{wang2021TrDiMP} } & CVPR21     &\ 39.9   &\ 34.8   &\ 48.7  &\ 26.3   &\ 26   \\ 
06    &  \textbf{ STARK~\cite{yan2021Stark}   }   &  ICCV21     &\ 44.5   &\ 39.6  &\ 55.7   &\ 28.1   &\ 42  \\ 
07    &  \textbf{ TransT~\cite{chen2021transt}   }   &  CVPR21     &\ 54.3  &\ 56.5  &\ 68.8   &\ 18.5   &\ 50  \\ 
08    &  \textbf{ ToMP50~\cite{mayer2022Tomp}  }   &  CVPR22   &\ 37.6   &\ 32.8   &\ 47.4   &\ 26.1   &\ 25  \\ 
09    &  \textbf{ OSTrack~\cite{ye2022Ostrack} }   &  ECCV22   &\ 55.4  &\ 60.4   &\ 71.1   &\ 92.1   &\ 105  \\
10    &  \textbf{ AiATrack~\cite{gao2022AIa}   }   &  ECCV22     &\ 57.4   &\ 59.7   &\ 72.8   &\ 15.8   &\ 38  \\ 
11    &  \textbf{ MixFormer~\cite{cui2022Mixformer}   }   & CVPR22     &\ 49.9   &\ 49.6   &\ 63.0   &\ 35.6   &\ 25  \\
12    &  \textbf{ SimTrack~\cite{chen2022SimTrack}   }   & ECCV22     &\ 55.4   &\ 57.5  &\ 69.9   &\ 57.8   &\ 40  \\ 
13    &  \textbf{ ROMTrack~\cite{cai2023robust} } &ICCV23  &\ 56.3  &\ 56.6 &\ 65.9 &\ 92.1   &\ 62 \\
14    &  \textbf{ GRM~\cite{gao2023generalized}} &CVPR23   &\ 57.8  &\ 62.8 &\ 72.4  &\ 99.8  &\ 45 \\
15    &  \textbf{ CiteTracker~\cite{li2023citetracker}}      &ICCV23   &\ 58.6   &\ \textbf{64.8}  &\ 73.2   &\ 244.5 &\ 12  \\ 
16    &  \textbf{ ARTrack~\cite{wei2023autoregressive} } &CVPR23  &\ 52.0  &\ 49.5 &\ 60.3 &\ 172.0   &\ 26  \\ 
17    &  \textbf{ AQATrack~\cite{xie2024autoregressive} }      &CVPR24  &\ \textbf{59.2}   &\ 63.7  &\ 74.3   &\ 78.5  &\  68 \\ 
18    &  \textbf{ ODTrack~\cite{zheng2024odtrack}}      &AAAI24  &\ 56.8   &\ 58.1  &\ 66.2   &\ 92.0   &\ 32  \\ 
19    &  \textbf{ EVPTrack~\cite{shi2024explicit} }      &AAAI24  &\ 58.4   &\ 62.8  &\ 73.9   &\ 73.0     &\ 71 \\ 
20    &  \textbf{ ARTrackV2~\cite{bai2024artrackv2} } &CVPR24   &\ 55.4 &\ 53.9 &\ 64.4 &\ 101.0    &\ 94  \\
21    &  \textbf{ LoRAT~\cite{lin2024tracking} } &ECCV24  &\ 57.8 &\ 56.7 &\ 66.0 &\ 99.0    &\ 209  \\
\hline
22    &  \textbf{ HDETrack~\cite{wang2024event}}      &CVPR24  &\ 57.8   &\ 62.2  &\ 73.5   &\ 92.1   &\ 105  \\ 
23    &  \textbf{ HDETrack V2}      &-  &\ 59.0  &\ 63.8 &\ \textbf{74.9}  &\ 92.1  &\ 35  \\ 
\hline \toprule [0.5 pt]
\end{tabular}
}
\end{table}

\noindent $\bullet$ \textbf{Results on FE240hz Dataset.\label{FE240HZ}}
As shown in Table~\ref{FE240table}, our baseline OSTrack achieves 60.0/89.7 on the SR/PR metric, meanwhile, HDETrack achieves 62.3/92.6 which is significantly better than the baseline method. After further improvement, our method demonstrated great advantages and reached new SOTA levels on SR and PR. Our tracker also beats other SOTA trackers including event-based trackers (e.g., STNet and EFE), and Transformer trackers (like TransT, STARK) by a large margin. These results fully validate the effectiveness of our proposed method for event-based tracking.

\begin{table}
\center
\small     
\caption{Experimental Results (SR/PR) on FE240hz Dataset.} 
\label{FE240table}
\resizebox{\columnwidth}{!}{
\begin{tabular}{cccccccccc}
\hline \toprule [0.5 pt]
\textbf{STNet}  &\textbf{TransT}  &\textbf{STARK}    &\textbf{PrDiMP}  &\textbf{EFE}    &\textbf{SiamFC++}    \\ 
58.5/89.6      &56.7/89.0        &55.4/83.7       &55.2/86.8        &55.0/83.5       &54.5/85.3 \\ 
\hline 
\textbf{DiMP}   &\textbf{ATOM}    &\textbf{Ocean}    &\textbf{OSTrack}    &\textbf{HDETrack}  &\textbf{Ours}     \\ 
53.4/88.2      &52.8/80.0         & 50.2/76.4        &57.1/89.3          &59.8/92.2    &60.4/93.2    \\         
\hline \toprule [0.5 pt]
\end{tabular}
}
\end{table}

\noindent $\bullet$ \textbf{Results on VisEvent Dataset.\label{VisEvent}} 
As shown in Table~\ref{Viseventtable}, we report the tracking results on the first large scale RGB-E dataset VisEvent, and compare our method with multiple recent strong trackers. Specifically, our baseline OSTrack~\cite{ye2022Ostrack} achieves $34.5, 50.1,$ and $41.6$ on SR, PR, and NPR, respectively, meanwhile, HDETrack is $37.3, 54.6,$ and $44.5$ on these metrics. By incorporating temporal knowledge distillation and optimization strategies during testing, our method has been further improved and achieved optimal performance in various metrics. These results demonstrate that our proposed method can enhance the event-based tracking results by learning from multi-modal input data. Compared with other Transformer based trackers, such as the STARK~\cite{yan2021Stark}, we can find that our results are much stronger than this tracker, with an improvement of +3.2 and +14 on SR and PR. We also beat the STNet and TransT on the PR metric, which fully validated the effectiveness of our proposed strategy for event-based tracking. 

\begin{table}
\center
\small     
\caption{Results on VisEvent Dataset. EF and MF are short for early fusion and middle-level feature fusion.} 
\label{Viseventtable}
\begin{tabular}{c|l|ccc}
\hline \toprule [0.5 pt] 
&\textbf{Trackers}   &\textbf{SR}  &\textbf{PR} &\textbf{NPR} \\   
\hline 
\multirow{9}{*}{\rotatebox{90}{\textbf{RGB + Event Input}}}
&\textbf{CEUTrack} &{64.89}   &{69.06} &{73.81} \\
&\textbf{LTMU (EF)} &  60.10  & 66.76 & 69.78\\
&\textbf{PrDiMP (EF)}&  57.20   & 64.47 & 67.02\\
&\textbf{CMT-MDNet (MF)} & 57.44   & 67.20  & 69.78\\
&\textbf{ATOM (EF) }  & 53.26 & 60.45 & 63.41\\
&\textbf{SiamRPN++ (EF) } & 54.11   & 60.58 & 64.72\\
&\textbf{SiamCAR (EF) } & 52.66 & 58.86  & 62.99\\
&\textbf{Ocean (EF) }  & 43.56  & 52.02 & 54.21\\
&\textbf{SuperDiMP (EF) } & 36.21   & 46.99 &42.84 \\
\hline   
\multirow{6}{*}{\rotatebox{90}{\textbf{Event Input}}}
&\textbf{STNet (Event-Only)}  & 39.7   & 49.2 &- \\
&\textbf{TransT (Event-Only)}  & 39.5   & 47.1  &- \\
&\textbf{STARK (Event-Only)}  & 34.8  & 41.8  &- \\
&\textbf{OSTrack (Event-Only)}  & 34.5    & 50.1   & 41.6 \\
&\textbf{HDETrack (Event-Only)}  &37.3 &54.6    &44.5   \\
&\textbf{HDETrack V2 (Event-Only)}  &38.0    &55.8    &45.1   \\
\hline \toprule [0.5 pt]
\end{tabular}
\end{table}

\noindent $\bullet$ \textbf{Results on FELT Dataset.\label{wang2024long}}
As shown in Table~\ref{FELT_benchmark_results}, we report our tracking results on the first long-term RGB-Event tracking dataset FELT. Note that, the compared baseline methods are re-trained on the training subset of FELT using their default settings and hyper-parameters to achieve a relatively fair comparison. It is easy to find that our baseline OSTrack achieves $37.4, 46.9, 44.0$ on the SR, PR, and NPR metrics, meanwhile, HDETrack obtains $39.0, 49.4, 45.9$ which are significantly better than theirs. Based on this, HDETrack V2 has surpassed its predecessor in SR, PR, and NPR, achieving $39.9$, $50.9$, and $47.4$, respectively. Furthermore, the new tracking results are also better than those of most of the compared trackers, including MixFormer, HIPTrack, GRM, etc. These experimental results fully demonstrate the effectiveness of our proposed hierarchical knowledge distillation from multi-modal to event-based tracking networks and the generalization ability brought by TTT and ASR.

\begin{table}[!htp]
\center
\small   
\caption{Tracking Results on the Long-term FELT SOT Dataset (Event-Only). } 
\label{FELT_benchmark_results}
\begin{tabular}{l|c|ccc}
\hline \toprule [0.5 pt]
\textbf{Trackers} & \textbf{Source}    &\textbf{SR}  &\textbf{PR}    &\textbf{NPR}  \\
\hline
\textbf{01. MixFormer  } \cite{cui2022Mixformer}       &CVPR22            & 38.9  &50.4  &44.4                 \\ 
\textbf{02. AiATrack  } \cite{gao2022AIa}       &ECCV22         &40.3 &51.6 &47.3                  \\ 
\textbf{03. STARK  }  \cite{yan2021Stark}      &ICCV21             &39.3 &50.8 &45.0                \\ 
\textbf{04. OSTrack  } \cite{ye2022Ostrack}      &ECCV22            &37.4 &46.9 &44.0                \\ 
\textbf{05. GRM  } \cite{gao2023generalized}      &CVPR23            &39.2 &48.9 &45.9               \\ 
\textbf{06. PrDiMP  } \cite{martin2020PrDimp}      &CVPR20         &34.9 &44.5 &41.6       \\
\textbf{07. DiMP50 } \cite{goutam2019Dimp}      &ICCV19             &37.8 &48.5 &45.5               \\ 
\textbf{08. SuperDiMP } \cite{danelljan2019pytracking}      &-         &37.8 &47.8 &44.2            \\ 
\textbf{09. TransT} \cite{chen2021transt}      &CVPR21          &35.2 &45.1 &42.4          \\ 
\textbf{10. TOMP50} \cite{mayer2022Tomp}      &CVPR22          &37.7 &49.2 &45.1                \\ 
\textbf{11. ATOM}  \cite{martin2019Atom}     &CVPR19               &22.3 &28.4 &26.7                 \\ 
\textbf{12. KYS}  \cite{bhat2022SKys}     &ECCV20              &22.5 &29.5 &26.1               \\  
\textbf{13. HIPTrack}  \cite{cai2024hiptrack}     &CVPR24             &37.5 &47.5 &43.9               \\  
\hline
\textbf{14. HDETrack\cite{wang2024event} }      &CVPR24       &39.0 &49.4 &45.9                \\
\textbf{15. HDETrack V2}      &-        &39.9 &50.9 &47.4              \\ 
\hline \toprule [0.5 pt]
\end{tabular}
\end{table}

\subsection{Ablation Study}~\label{Ablation Study}

\begin{table} 
\center
\small
\caption{Component Analysis on EventVOT Dataset.}
\label{CAResults}
\begin{tabular}{c|cccc|c}
\hline \toprule [0.5 pt]
\textbf{No.} & \textbf{Base} & \textbf{SKD} & \textbf{FKD} & \textbf{RKD}  & \textbf{EventVOT} \\
\hline
1 &\cmark   &          &         &                  &\ 55.4/60.4/71.1        \\
2 &\cmark   &\cmark    &         &                  &\ 56.5/60.8/72.4         \\
3 &\cmark   &    &\cmark         &                  &\ 56.4/60.4/72.2          \\
4 &\cmark   &   &    &\cmark                        &\ 56.2/60.6/71.9           \\
\hline 
5 &\cmark   &\cmark   &\cmark    &                  &\ 57.2/61.3/73.0        \\
6 &\cmark   &\cmark   &    &\cmark                  &\ 57.5/62.2/73.1         \\
7 &\cmark   &   &\cmark    &\cmark                  &\ 57.6/62.1/73.3          \\
8 &\cmark   &\cmark    &\cmark   &\cmark            &\ 57.8/62.2/73.5           \\  
\hline \toprule [0.5 pt]
\end{tabular}
\end{table}

\begin{table} 
\center
\caption{Analysis of Newly Added Modules on EventVOT Dataset.}
\label{CAResults2}
\begin{tabular}{c|cccc|cc}
\hline \toprule [0.5 pt]
\textbf{No.} & \textbf{HDETrack} & \textbf{TFT} & \textbf{TTT} & \textbf{ASR}  & \textbf{SR / PR / NPR} \\
\hline
1 &\cmark   &          &          &                     &\ 57.8/62.2/73.5         \\
2 &\cmark   &\cmark    &          &                     &\ 58.4/62.9/74.1          \\ 
3 &\cmark   &\cmark    &\cmark    &                     &\ 58.6/63.4/74.4           \\
4 &\cmark   &\cmark    &\cmark    &\cmark               &\ 59.0/63.8/74.9            \\
\hline \toprule [0.5 pt]
\end{tabular}
\end{table}

\begin{figure*}[!htp]
\center
\includegraphics[width=\linewidth]{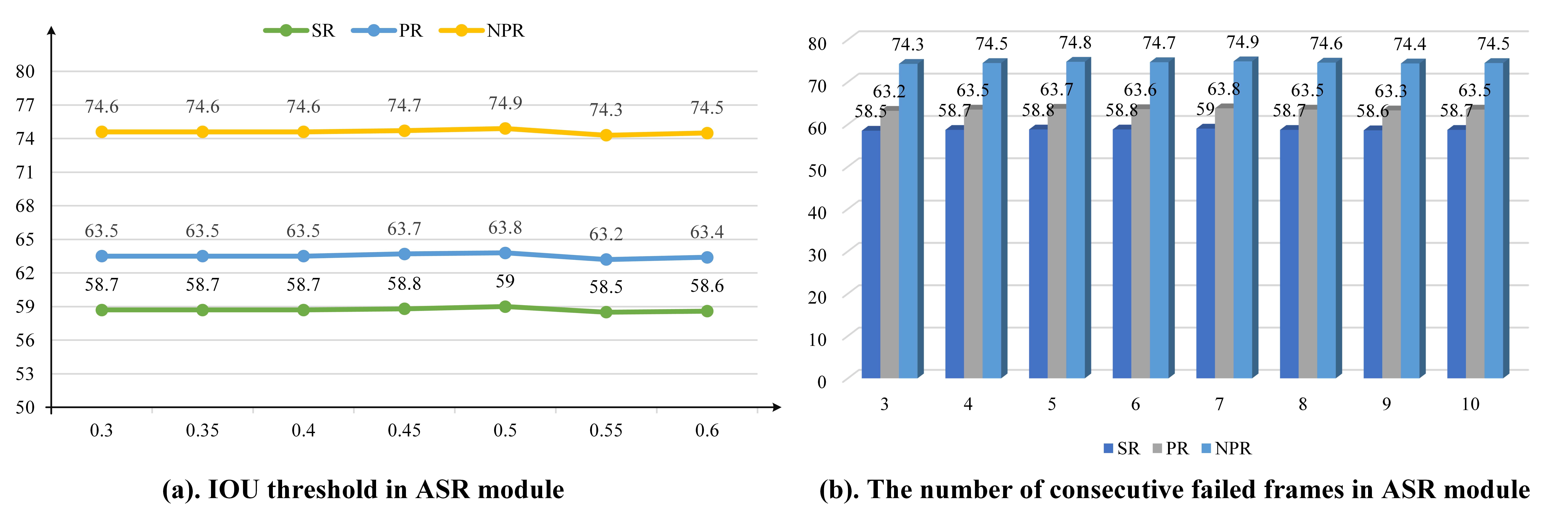}
\caption{(a). IOU threshold in ASR module;
(b). The number of consecutive failed frames in the ASR module.} 
\label{ablation2}
\end{figure*}

\noindent $\bullet$ \textbf{Analysis on Hierarchical Knowledge Distillation.}
Following HDETrack, for the EventVOT dataset, we stack the event stream into event images and event voxels and apply hierarchical knowledge distillation based on multi-view settings. As shown in Table~\ref{CAResults}, the base denotes the tracker trained solely with three tracking loss functions, similar to the approach used by OSTrack. This base model achieves $55.4/60.4/71.1$ on the EventVOT datasets, respectively. The introduction of new distillation losses, such as similarity-based, feature-based, and response-based knowledge distillation functions, all improve the performances. It is noteworthy that the tracking performance is further enhanced in multi-view settings when all these distillation strategies are combined. Based on these experiments, we conclude that the proposed hierarchical knowledge distillation strategies significantly contribute to event-based tracking.

\noindent $\bullet$ \textbf{Analysis of the Newly Proposed TFT based KD, TTT, and ASR.}
To help the model learn more temporal relationships between video frames, we adjust the input strategy: each template frame now corresponds to multiple search frames during the training stage. Additionally, we propose the Temporal Fourier Transform (TFT) to establish a temporal relationship distillation among multiple tracking results, as shown in Table~\ref{CAResults2}. Clearly, as the model learns more inter-frame temporal relationships, its performance improves. On the EventVOT dataset, the SR, PR, and NPR all showed enhancements to a certain extent, achieving $58.4/62.9/74.1$.
Subsequently, we introduce Test Time Tuning (TTT) into the tracking process to reduce the distribution gap between the training and test sets, thereby improving the model's generalization of the test set. 

To alleviate the issue of tracking failures caused by fast-moving targets or targets moving out-of-view, we introduce an Adaptive Search Region (ASR) strategy, allowing the model to expand the search region to find reliable tracking targets in such cases. The results show that our efforts during inference were worthwhile, with the model achieving $59.0/63.8/74.9$ of SR, PR, and NPR on the EventVOT dataset, respectively. Ultimately, compared to the previous version (HDETrack), our model shows an accuracy improvement of $+1.2, +1.6, +1.4$ on each metric. These analytical experimental results validate the rationality of our improvements, bringing substantial performance gains to event tracking.

\begin{table}
\center
\small     
\caption{Ablation Studies of Parameter Analysis on EventVOT Dataset.} 
\label{event_repr}
\begin{tabular}{l|lll}
\hline \toprule [0.5 pt]
\textbf{Optimized modules}    &\textbf{SR}   & \textbf{PR}  & \textbf{NPR}  \\
\hline
\text{1. MLP }     &58.0   &62.5  &73.7  \\
\text{2. MLP + attn.proj }     &58.3    &62.9   &74.1  \\
\text{3. MLP + attn.proj + attn.qkv}     &\textbf{58.4}   &\textbf{63.3}  &\textbf{74.3}  \\
\hline
\textbf{Expansion factor $\theta$}    &\textbf{SR}   & \textbf{PR}  & \textbf{NPR}  \\
\hline
\text{1. $\theta = 1.1$ }     &58.5   &63.3  &74.4  \\
\text{2. $\theta = 1.3$ }     &58.7   &63.5  &74.6  \\
\text{3. $\theta = 1.5$ }     &\textbf{59.0}   &\textbf{63.8}  &\textbf{74.9}  \\
\text{4. $\theta = 1.7$ }     &58.8   &63.6  &74.6  \\
\hline \toprule [0.5 pt]
\end{tabular}
\end{table}

\begin{figure}
\center
\includegraphics[width=3.3in]{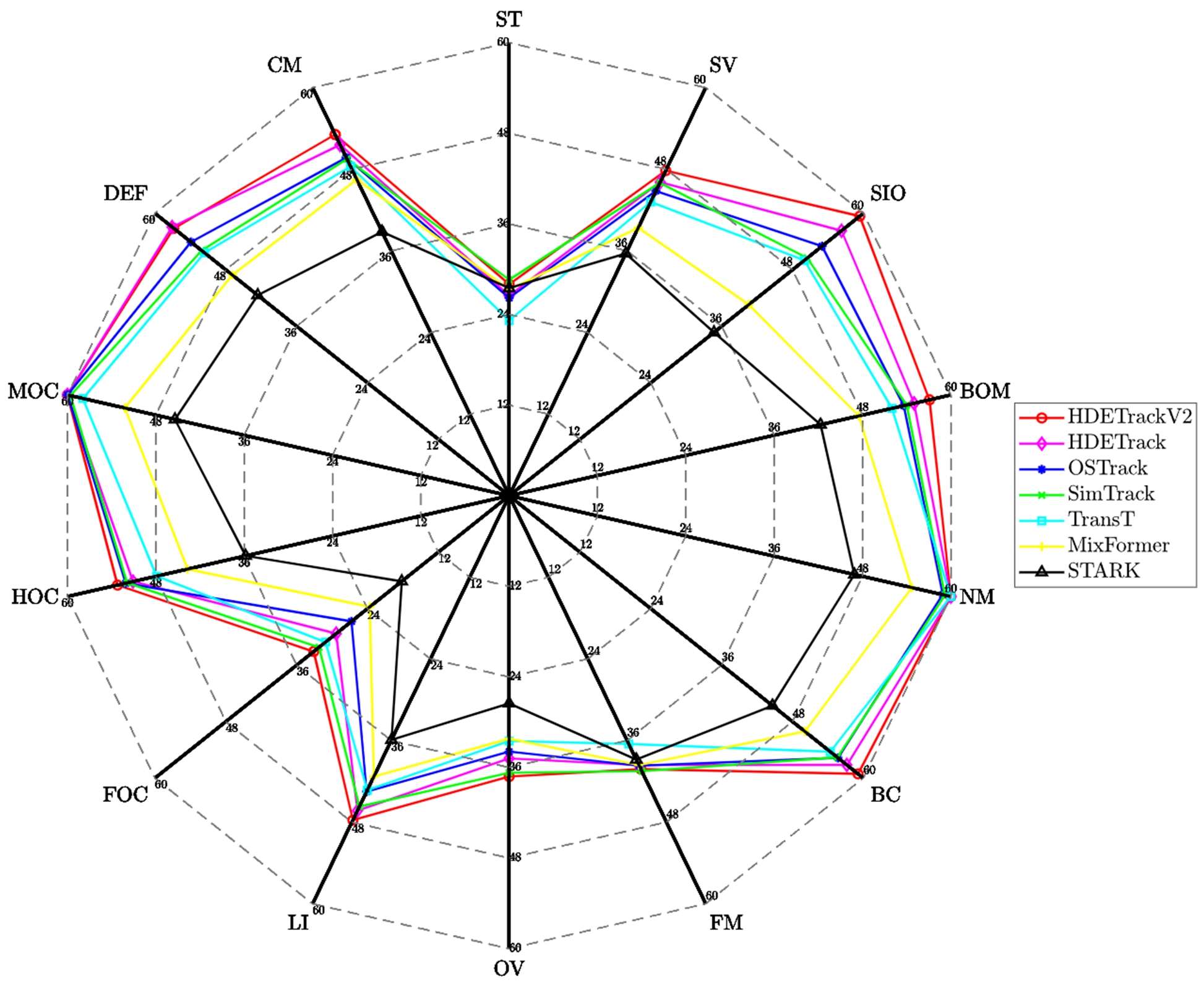}
\caption{Tracking results (SR) under each challenging factor.} 
\label{attributeResults}
\end{figure}

\noindent $\bullet$ \textbf{Analysis on Tracking in Specific Challenging Environments.~} 
As shown in Table~\ref{AttributeList}, our proposed EventVOT dataset reflects 14 core challenging factors in the tracking task. To demonstrate the advantages of our tracker in challenging scenarios, we also report the results of our tracker and other state-of-the-art trackers under each challenging attribute in Fig.~\ref{attributeResults}. It can be seen that HDETrack V2 exhibits superior performance in handling challenging attributes such as Deformation (DEF), Camera motion (CM), Similar interfering objects (SIO), and Background clutter (BC), among others. Additionally, it also achieves comparable tracking results across various other attributes, thereby validating that our hierarchical knowledge distillation strategy works well for transferring knowledge from multi-modal/multi-view data. Simultaneously, in the challenging scenarios of Fast motion (FM) and Out-of-view (OV), our newly introduced Test Time Tuning strategy and Adaptive Search Region have proven to be more effective for event-based object tracking compared to HDETrack.

\noindent $\bullet$ \textbf{Analysis on Test Time Tuning Strategy.~}
Next, we will discuss the detailed ablation studies on the TTT strategy. Firstly, we analyze the number of initial frames used for training in TTT on the EventVOT dataset. Our rationale is that in the initial frames of a video, the target's movement is typically minimal, allowing the tracker to achieve good results on these initial frames. Thus, we select the first $n$ frames in a video as the training samples, using their tracking results from the tracker as the pseudo-labels for self-supervised learning. As shown in Fig.~\ref{ablation1} (a), setting the number of initial training frames $n$ to $5$ yields the best results.
Additionally, we analyze the impact of the parameters $r$ and $alpha$ in LoRA. Inspired by the efficiency of the LoRA mechanism for fine-tuning large models, we incorporate LoRA into TTT to optimize the model in the inference stage. Fig.~\ref{ablation1} (b) shows that the different parameter settings have varying effects on the model performance. In this study, we set $r$ to $16$ and $alpha$ to $32$, which provide relatively optimal performance. 
Subsequently, we review the optimizable network layers of LoRA in TTT. As shown in Table~\ref{event_repr}, the best results are achieved when 'mlp', 'attn.proj', and 'attn.qkv' are all used, this experiment indicates that these linear layers are indispensable. 
Overall, by adjusting various parameters in the TTT strategy, we can improve the model's generalization ability on the test set.

\noindent $\bullet$ \textbf{Analysis on Adaptive Search Region Strategy.~}
In conventional tracking tasks, the search regions are typically set based on the bounding box from the previous frame’s tracking result, adjusted by a suitable search factor. However, in highly challenging scenarios, such as target drift, the tracker may fail if it cannot locate the intended target. To address this issue, we simply expand the search region appropriately. Here, we analyze when to expand the search region and by what factor for optimal results on the EventVOT dataset. As shown in Fig.~\ref{ablation2}, we set an IoU threshold, $\tau$, to detect potential target drift. If the IoU between two successive frames falls below $\tau$, drift may occur; If the IoU remains below $\tau$ for $k$ consecutive frames, we conclude that the tracker has likely failed due to target drift. At this point, we expand the ratio of crop size of the search region in the next frame to recapture the target within the tracking view.
Consequently, to study the effect of the expansion factor on tracking performance, we further analyze the optimal expansion factor $\theta$ of the search region in Table~\ref{event_repr}. Finally, we set $\tau$, $k$, $\theta$ to $0.5$, $7$, and $1.5$, respectively, and achieve the best performance in tracking.

\begin{figure*} 
\center
\includegraphics[width=7in]{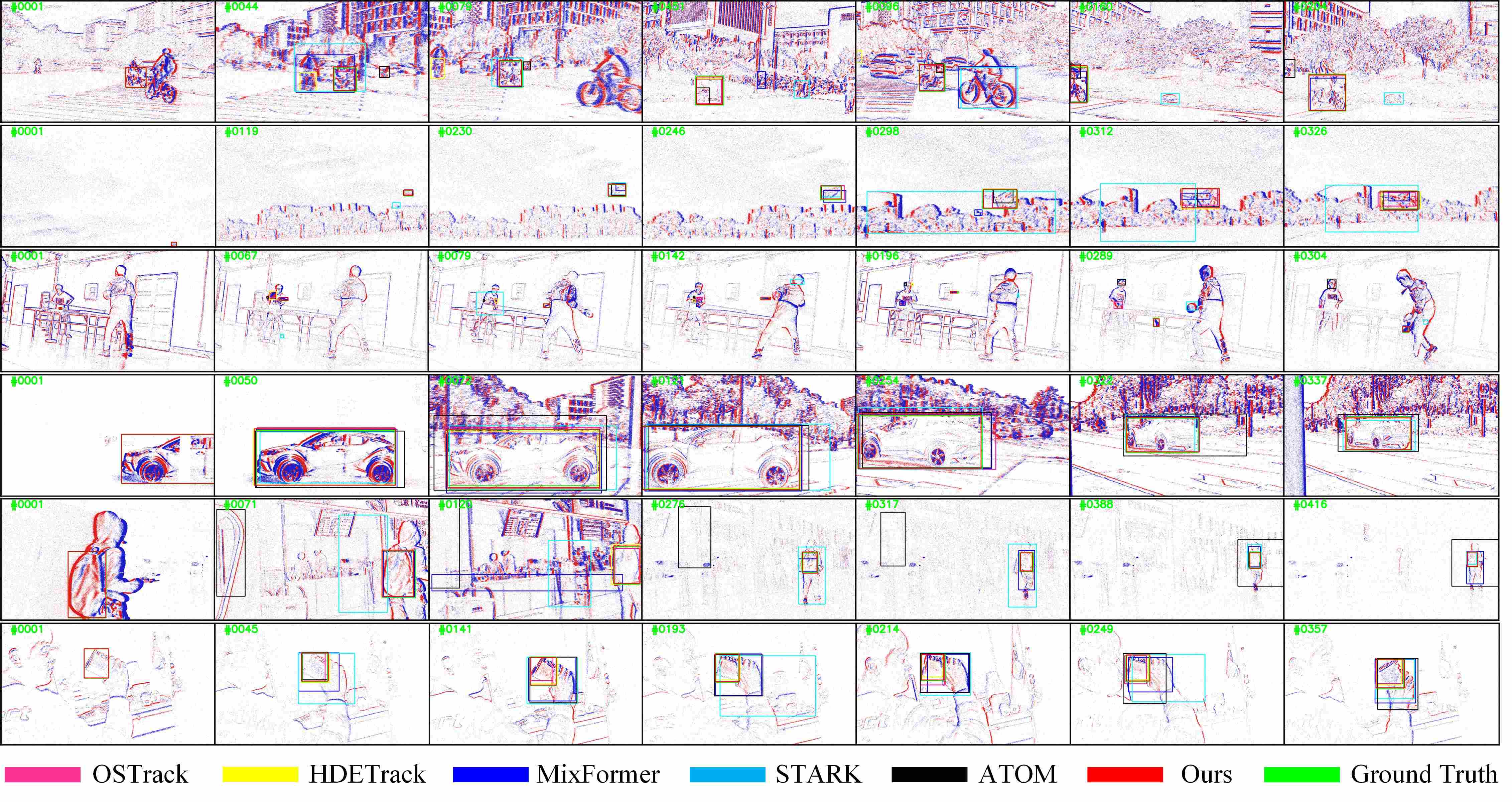}
\caption{Visualization of the tracking results of ours and other SOTA trackers.}  
\label{trackingResults}
\end{figure*}

\begin{figure}
\center
\includegraphics[width=3.5in]{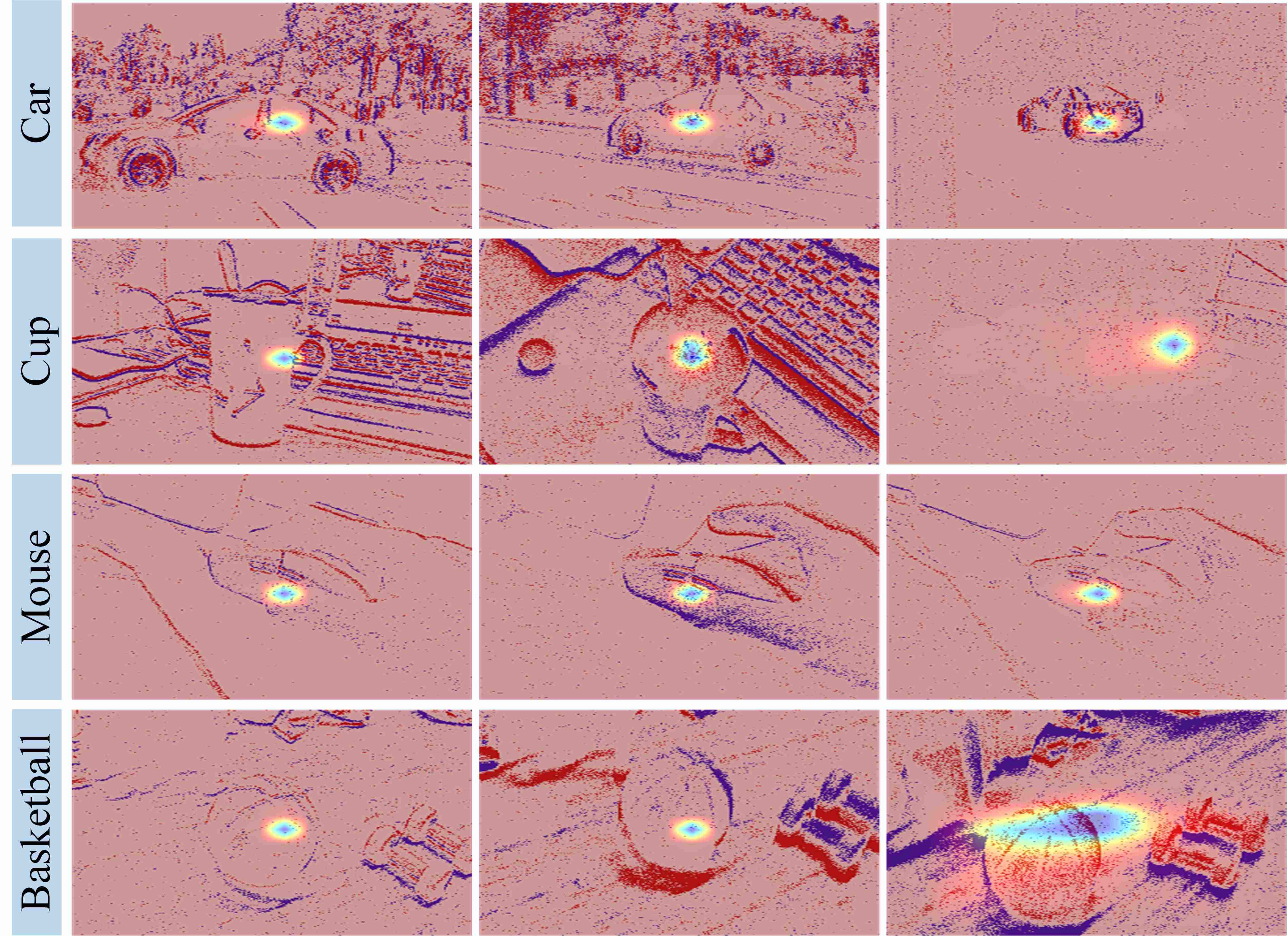}
\caption{Response maps predicted by our HDETrack V2.}  
\label{response_map}
\end{figure}

\begin{figure}
\center
\includegraphics[width=3.5in]{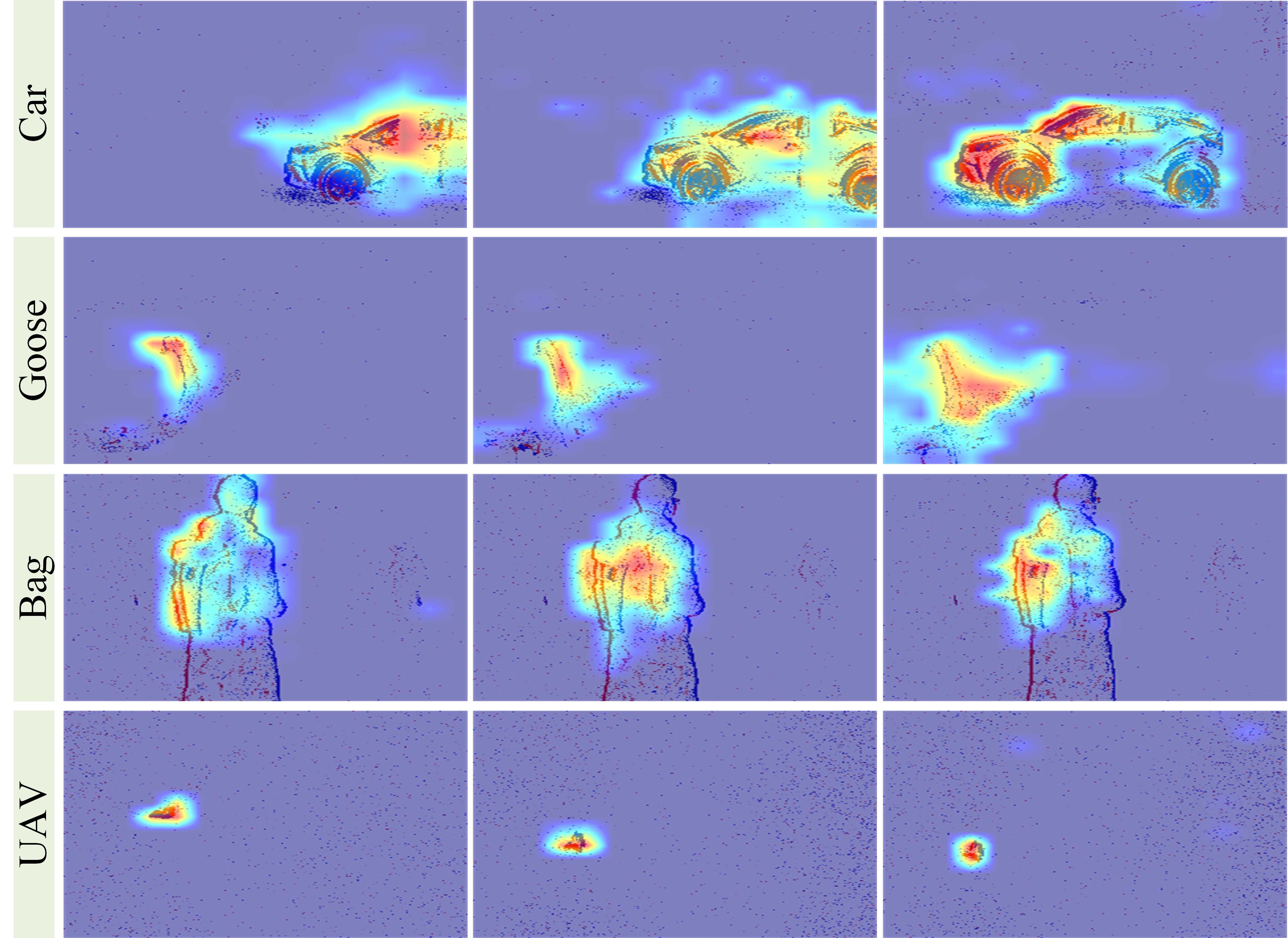}
\caption{Attention maps predicted by our HDETrack V2.}  
\label{attention_map}
\end{figure}

\subsection{Visualization}~\label{Visualization}

\noindent $\bullet$ \textbf{Visual Comparison of Metric Results and Tracking Results.~}
Beyond the quantitative analysis previously discussed, we have also performed some visualizations of the proposed tracking algorithm to offer readers a more intuitive and comprehensive understanding of the operation and effectiveness of our tracking framework. 
As shown in Fig.~\ref{PRSRNPRfig}, we visualize the SR, PR, and NPR metrics of various benchmarks on our proposed EventVOT dataset. It is evident that our proposed HDETrack V2 achieves optimal performance on SR and NPR compared to numerous baselines. Our method is a close runner-up to CiteTracker in terms of PR metric. Through this comparative visualization, the superiority of our proposed method can be more clearly seen. 
Besides, we also visualize the tracking results of ours and other SOTA trackers in Fig.~\ref{trackingResults}, including OSTrack, ATOM, MixFormer, STARK, and HDETrack. We can observe that tracking with an event camera presents both interesting and challenging aspects. While these trackers perform satisfactorily in straightforward scenarios, there remains substantial scope for enhancement.

\noindent $\bullet$ \textbf{Visualization of the Response Maps and Attention Maps.~} 
To better understand the extent to which our model attends to the target, we visualized both the response maps and attention maps of the Transformer network during tracking. As shown in Fig.~\ref{response_map}, the highlighted region represents the area with the highest response to the tracking result within the entire image. 
Notably, the response area of the model aligns closely with the target object’s search region. Meanwhile, we visualized the attention map of the Transformer network in Fig.~\ref{attention_map}, where it is evident that the network precisely focuses its attention on the target's location. These visualizations provide a clearer insight into the effectiveness of our network and demonstrate its potential for more accurate visual object tracking tasks.

\subsection{Limitation Analysis}
Although our work has achieved decent accuracy, there are still some areas that can be further improved:
Firstly, the conversion from the Event stream to a fixed-frame video, which adapts well to existing tracking frameworks for evaluation, maybe a worthwhile research direction for Event tracking in terms of dense video annotation and high-frame-rate tracking.
Secondly, various challenging factors were not utilized during the training phase. The model is unaware of the nature of the challenges it encounters, making it difficult to handle these difficulties effectively. Merely relying on feature engineering cannot adequately address these issues in the short term. We consider introducing large language models in our future work to enable the model to understand various challenging factors and design automatic strategies to cope with them, fundamentally solving these problems and achieving better tracking results.

\section{Conclusions and Future Works}~\label{sec:Conclusions} 
In this paper, we present HDETrack V2, a hierarchical knowledge distillation framework for event-based tracking. It facilitates the transfer of multi-modal/multi-view knowledge to an unimodal tracker through carefully designed strategies, including similarity-based, feature-based, and response map-based knowledge distillation. Compared with our previous version, the Temporal Fourier Transform is used to establish the temporal relationships between video frames to enhance the knowledge distillation process. 
Furthermore, attributing to the Test Time Tuning and Adaptive Search Region strategies, the model can perform better during the inference phase. 
To bridge the data gap, a large-scale high-resolution event-based tracking dataset has been proposed, termed EventVOT. 
Extensive evaluations on multiple tracking benchmarks demonstrate that our tracker obtains a notable performance over other prevailing trackers. 
In our future works, we will consider developing a lightweight network capable of handling high-resolution and more challenging tracking scenarios, further broadening the applicability of event-based tracking algorithms.

\small{ 
\bibliographystyle{IEEEtran}
\bibliography{reference}
}

\end{CJK}
\end{document}